\documentclass[lettersize,journal]{IEEEtran}
\usepackage{amsmath,amsfonts}
\usepackage{algorithmic}
\usepackage{algorithm}
\usepackage{array}
\usepackage{textcomp}
\usepackage{stfloats}
\usepackage{url}
\usepackage{verbatim}
\usepackage{graphicx}
\usepackage{epstopdf}
\usepackage{cite}
\usepackage{makecell}
\usepackage[colorlinks,
linkcolor=green,
anchorcolor=green,
citecolor=green]{hyperref}
\usepackage{framed,multirow,rotating,float}
\usepackage[table]{xcolor}

\usepackage{amssymb}
\usepackage{latexsym}
\usepackage{booktabs}
\usepackage{amsmath}
\usepackage{gensymb}
\usepackage{pifont}
\usepackage{bbding}

\usepackage{subfigure}

\begin{document}

\title{Rethinking Object Saliency Ranking: A Novel Whole-flow Processing Paradigm}

\author{\IEEEauthorblockN{Mengke Song$^{1}$,~Linfeng Li$^{4}$,~Dunquan Wu$^{4}$,~Wenfeng Song$^{5}$,~Chenglizhao Chen$^{1,2,3*}$\thanks{* Corresponding author: Chenglizhao Chen, cclz123@163.com.}\\
	\IEEEauthorblockA{$^{1}$China University of Petroleum (East China)}\\
$^{2}$Shandong Provincial Key Laboratory for Distributed Computer Software Novel Technology\\
$^{3}$Jiangsu Key Laboratory of Image and Video Understanding for Social Safety\\
        $^{4}$Qingdao University~~~~$^{5}$Beijing Information Science and Technology University\vspace{-0.8cm}}}

\markboth{IEEE TRANSACTIONS ON IMAGE PROCESSING, VOL.XX, NO.XX, XXX.XXXX}%
{Shell \MakeLowercase{\textit{et al.}}: A Sample Article Using IEEEtran.cls for IEEE Journals}


\maketitle

\begin{abstract}
Existing salient object detection methods are capable of predicting binary maps that highlight visually salient regions. However, these methods are limited in their ability to differentiate the relative importance of multiple objects and the relationships among them, which can lead to errors and reduced accuracy in downstream tasks that depend on the relative importance of multiple objects.
To conquer, this paper proposes a new paradigm for saliency ranking, which aims to completely focus on ranking salient objects by their ``importance order''. While previous works have shown promising performance, they still face ill-posed problems.
First, the saliency ranking ground truth (GT) orders generation methods are unreasonable since determining the correct ranking order is not well-defined, resulting in false alarms.
Second, training a ranking model remains challenging because most saliency ranking methods follow the multi-task paradigm, leading to conflicts and trade-offs among different tasks.
Third, existing regression-based saliency ranking methods are complex for saliency ranking models due to their reliance on instance mask-based saliency ranking orders. These methods require a significant amount of data to perform accurately and can be challenging to implement effectively.
To solve these problems, this paper conducts an in-depth analysis of the causes and proposes a whole-flow processing paradigm of saliency ranking task from the perspective of ``GT data generation'', ``network structure design'' and ``training protocol". The proposed approach outperforms existing state-of-the-art methods on the widely-used SALICON set, as demonstrated by extensive experiments with fair and reasonable comparisons.
The saliency ranking task is still in its infancy, and our proposed unified framework can serve as a fundamental strategy to guide future work.
The code and data will be available at \emph{\url{https://github.com/MengkeSong/Saliency-Ranking-Paradigm}}.
\end{abstract}


\section{Introduction}
\label{intro}
\begin{figure}[!t]
	\centering{\includegraphics[width=1\linewidth]{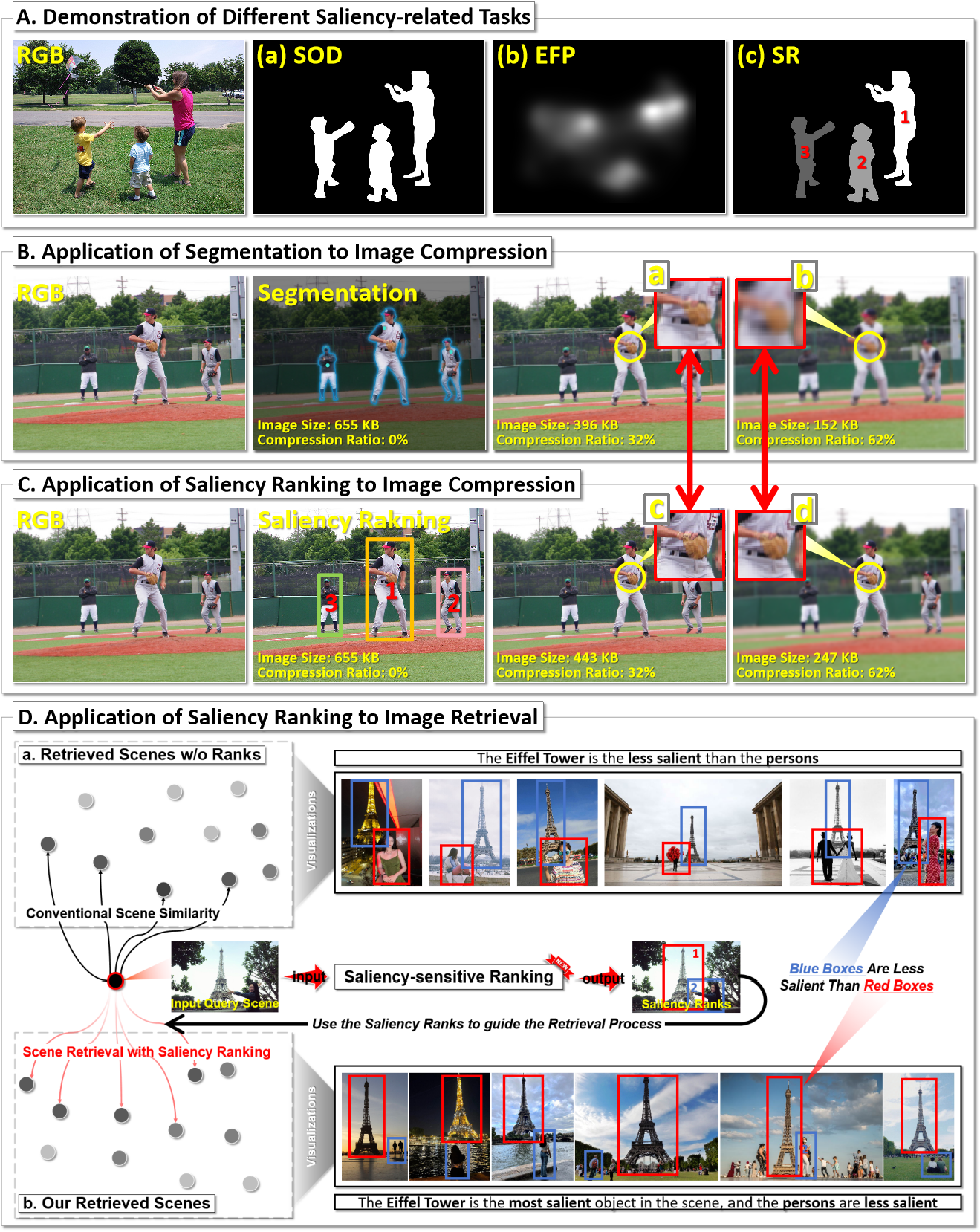}}
	\vspace{-0.7cm}
	\caption{Three saliency related tasks (A) regarding \textbf{s}alient \textbf{o}bject \textbf{d}etection (SOD), \textbf{e}ye \textbf{f}iaxtion \textbf{p}rediction (EFP), and \textbf{s}aliency \textbf{r}anking (SR). \textbf{B}, \textbf{C} and \textbf{D} provide a practical application of saliency ranking to image compression and image retrieval.
	Segmentation-based image compression (\textbf{B}) reveal that each object's importance or saliency degree is disregarded during the segmentation task. Consequently, when applying image compression, all objects undergo the same compression effect, leading to a certain loss of information in the more salient objects. Saliency ranking-based image compression (\textbf{C}) ensures that the compression effect on objects is inversely proportional to their saliency, resulting in higher preservation of information for more salient objects (comparing subfigure \textbf{B}-a and \textbf{C}-c, or \textbf{B}-b and \textbf{C}-d, in red boxes). \textbf{D}-a: existing semantically similar-based image retrieval results; \textbf{D}-b: our saliency-sensitive ranking-based image retrieval results. With the assistance of saliency ranking, our method can retrieve images based on saliency-sensitive rankings, making the retrieval results more fine-grained and accurate.}
	\label{fig:3tasks}
	\vspace{-0.65cm}
\end{figure}

Saliency detection is a fundamental task in computer vision. Previous research primarily focuses on \textbf{s}alient \textbf{o}bject \textbf{d}etection (SOD) task~\cite{Song-TIP22-MAD} and \textbf{e}ye \textbf{f}ixation \textbf{p}rediction (EFP) task~\cite{Han-TCYB16-Two}. SOD defines saliency in an absolute way using binary object-level pixel-wise saliency maps (Fig.~\ref{fig:3tasks}-A(a)), while EFP aims to predict scattered human eye fixations (Fig.~\ref{fig:3tasks}-A(b)). However, existing models for these tasks can only learn to detect all salient objects in an image equally without explicitly differentiating their different degrees of saliency (importance).

Therefore, \textbf{s}aliency \textbf{r}anking (SR) task is proposed to predict relative saliency of objects in a scene (Fig.~\ref{fig:3tasks}-A(c)), which enables us to clearly distinguish which object is more salient and know the relative importance among objects, thus benefiting downstream tasks, such as image compression~\cite{image-comp} (Fig.~\ref{fig:3tasks}-B, C) and image retrieval~\cite{image-Retrieval} (Fig.~\ref{fig:3tasks}-D).
Unlike SOD, SR is more fine-grained since it assigns unique relative saliency (importance) ranking orders to all salient objects based on their visual saliency and differentiates them by their saliency (importance) degree. Therefore, it can provide more detailed visual information regarding the interrelationships of salient objects rather than simply detecting them.

\begin{figure}[!t]
	\centering
	\includegraphics[width=1.0\linewidth]{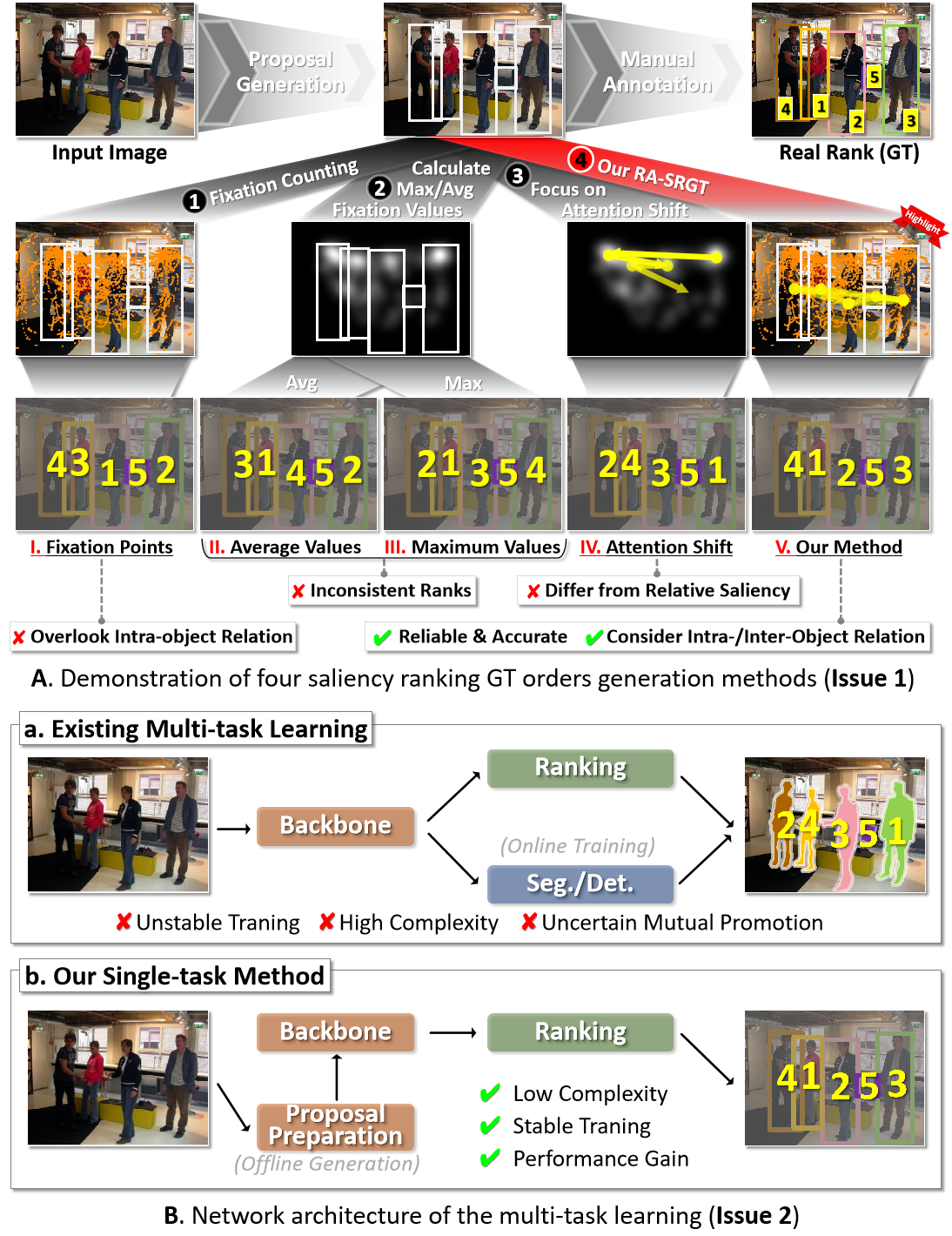}
	\vspace{-0.6cm}
	\caption{In subfigure \textbf{A}, Marks {\large{\textcircled{\small{1}}}} $\to$ {\large{\textcircled{\small{4}}}} denote fixation points-based, fixation maps-based, attention shift-based and our novel \textbf{r}elationship-\textbf{a}ware \textbf{s}aliency \textbf{r}anking \textbf{GT} orders generation method (RA-SRGT), respectively. Compared with other existing methods, our proposed RA-SRGT boosts the reliability and accuracy of saliency ranking GT orders generation. Subfigure \textbf{B} is the network architecture of the multi-task learning in saliency ranking task (a) and our proposed single-task method (b) (\textbf{Issue 2}).}
	\vspace{-0.5cm}
	\label{fig:motivation1}
\end{figure}

Despite the promising performance of existing SR methods~\cite{Islam-CVPR18-ranking,Siris-CVPR20-raning,Liu-TPAMI21-GCN}, they still encounter several ill-posed issues that may result in biased and inaccurate results. We can categorize these issues into three types:

\textbf{Issue 1:} \textit{The \textbf{g}round-\textbf{t}ruths (GT) of the ranking orders are improperly generated} (Fig.~\ref{fig:motivation1}-A).
Existing saliency ranking GT orders generation methods are mainly fixation points-based (Mark {\large{\textcircled{\small{1}}}}), fixation maps-based (Mark {\large{\textcircled{\small{2}}}}) and attention shift-based (Mark {\large{\textcircled{\small{3}}}}).
Among them, fixation points-based methods~\cite{Lv-CVPR21-camouflaged} assign the saliency ranking GT orders by calculating the number of fixation points within an object, while fixation maps-based methods~\cite{Liu-TPAMI21-GCN,Tian-CVPR22-BI} count the average/maximum pixel values from the fixation map of an object, since the fixation maps, displaying the distribution and density of multiple fixation points across a visual stimulus, are smoother than the fixation points which represents discrete locations where the eyes fixate. Besides, attention shift-based methods~\cite{Siris-CVPR20-raning,Kalash-TPAMI21,Fang-ICCV21-PPA} assign descending saliency scores to objects based on the order of ``attention shift\footnote{Attention shift order refers to the process of observers gazing from one object to another, which can reflect their interest and preference for the image content. The final saliency rank is an average across the saliency rank orders of multiple observers.}'' of observers over different objects in an image. We will detail these three methods in the following from the perspective of ``GT data generation''.


1) Fixation points-based saliency ranking methods rely on the discrete points of eye fixations within the objects acquired from human observers to assign object saliency ranking orders, which is implemented by counting the total number of fixation points within each object and assigning a higher saliency ranking order to objects with more fixation points. However, this naive saliency ranking order assignment may lead to inaccurate results. The reason is that these approaches only consider intra-object relationships, meaning that they focus on isolated objects without considering inter-object relationships. This limitation prevents fixation points-based methods from accurately determining how objects relate to each other or the overall composition of the image.
For example, Mark {\large{\textcircled{\small{1}}}} in Fig.~\ref{fig:motivation1}-A assigning GT saliency ranking orders based on fixation points within an object leads to the wrong order.

2) Fixation maps-based saliency ranking methods use fixation maps to assign object saliency ranking orders. These methods calculate the maximum or average pixel values within each object and assign a higher saliency ranking order to objects with larger pixel values. However, these methods suffer from inconsistent ranking results across different saliency ranking GT orders generation methods and are sensitive to noise occurring in fixation maps. For example, in Fig.~\ref{fig:motivation1}-A (II, III), when calculating the average or maximum pixel values from the fixation map (Mark {\large{\textcircled{\small{2}}}}), indistinct objects may be deemed as more salient.


3) Regarding attention shift-based methods (Mark {\large{\textcircled{\small{3}}}}), it has been noted by~\cite{Liu-TPAMI21-GCN} that the degree of an object's saliency mainly depends on the gaze duration (Fig.~\ref{fig:motivation1}-A (IV)), \emph{i.e.}, how humans sequentially select and shift attention from one object to another. It mainly relies on binary saliency prediction to rank objects rather than truly simulating the human attention shift process. Thus, attention shift differs mostly from relative saliency and is very close to the scan path prediction \cite{Sun-TPAMI21-Scanpath}.

It is important to note that all three types of saliency ranking GT orders generation methods can be problematic and result in inferior outcomes, adversely affecting model training and evaluation. However, another issue regarding ``network structure design'' arises in the saliency ranking task.


\textbf{Issue 2:} \textit{The effect of multi-task learning in saliency ranking and segmentation or detection is still an open question.}
Most existing saliency ranking methods adopt multi-task learning (``segmentation + ranking'' or ``object detection + ranking''). However, these combinations of tasks differ greatly in terms of datasets, models, and evaluation metrics.
In particular, existing object detection methods already perform well; for the saliency ranking task, multi-task learning (Fig.~\ref {fig:motivation1}-B (a)) could bring few benefits. Instead, it increases the complexity of the model (See Table~\ref{time0} for detailed proof) and leads to unstable training. Furthermore, the performance of multitask is not necessarily mutually enhance each other. Detailed analysis of the issue is presented in Sec.~\ref{issue2}.
Apart from the issue of multi-task learning, there exists another issue. We will detail it in the following.

%

%

\textbf{Issue 3:} \textit{Regression-based saliency ranking methods are unsuitable for ranking models}.
Existing saliency ranking methods rely heavily on regression-based instance segmentation, which is pixel-wise and requires a large quantity of data to achieve accurate results. These approaches can be quite complex and difficult to implement effectively.

Targeting the abovementioned issues in existing saliency ranking methods, this paper proposes a whole-flow processing paradigm of saliency ranking task, \emph{i.e.}, ``\textbf{GT data generation}'' (\textbf{Issue 1}) $\to $ ``\textbf{network structure design}'' (\textbf{Issue 2}) $\to $ ``\textbf{training protocol}'' (\textbf{Issue 3}), which could serve as a new fundamental strategy for future saliency ranking work.

To address \textbf{Issue 1}, we propose a novel \textbf{r}elationship-\textbf{a}ware \textbf{s}aliency \textbf{r}anking \textbf{GT} orders generation method (RA-SRGT, Mark {\large{\textcircled{\small{4}}}} in Fig.~\ref{fig:motivation1}-A). This is designed to overcome the limitations of existing saliency ranking GT orders generation methods, which produce inconsistent and varied results that do not align with the principles of the human visual attention system (Fig.\ref{fig:motivation1}-A (V)).
Compared to existing methods, our new GT orders generated by RA-SRGT are more precise and rational, and can be derived directly from fixation points without requiring re-labeling.

Regarding \textbf{Issue 2}, we offer a simple yet efficient solution for saliency ranking tasks that avoids the need for complicated multi-task learning approaches. We avoid complex training while retaining high detection accuracy by leveraging already high-performance and pre-trained object detection models to obtain object proposals offline. Our adaptable approach makes it a valuable alternative for real-world applications.


To solve the \textbf{Issue 3}, we propose to utilize pure object-wise classification to predict saliency ranking orders. This approach has several advantages over pixel-wise instance segmentation-based regression, since regarding data amount, object proposals are sparser than pixel-wise instance masks, which are better suited for classification than regression.
However, the flexible object input issue can arise with classification-based methods, which we address with our newly-devised \textbf{a}daptive \textbf{c}irculative \textbf{b}agging (ACB, Sec.~\ref{sec:ACB}), allowing for varying number of input proposals generated by any object detection model and making it more applicable to real-world scenarios.

The conducted experiments have demonstrated the superiority of the proposed approach when compared to state-of-the-art methods. The obtained results highlight the effectiveness of the unified framework in accurately ranking salient objects. 
%



To sum up, the main contributions of this work include:
\begin{itemize}
	\item We have pioneeringly conducted a thorough analysis of the three fundamental issues that commonly occur in existing saliency ranking methods and proposed a novel whole-flow processing paradigm for saliency ranking;
	\item We propose a brand-new saliency ranking GT orders generation method which exhibits greater conformity with the real ranking orders (GT data generation);
	\item We propose a general solution to the challenges of multi-task learning in saliency ranking. The proposed method is a novel proposal-based single-task approach that uses offline object proposals and pure classification to accurately rank objects and reflect human visual perception, which offers unparalleled simplicity, efficiency, and reliability (network structure design);
	\item We present a novel adaptive circulative processing approach that can handle varying numbers of input proposals from any object detection model (training protocol);
\end{itemize}

\section{Related Work}
\subsection{Salient Object Detection}
Previous \textbf{s}alient \textbf{o}bject \textbf{d}etection (SOD) works have mainly used handcrafted features (\emph{e.g.}, color contrast~\cite{Cheng-TPAMI15-Glo}, background prior~\cite{Wei-ECCV12-Geo}, and fixations~\cite{ZliuSH}) to detect salient objects, which limits their generalization ability and performance in complex scenarios. However, the emergence of deep learning has led to the development of CNN-based SOD~\cite{Siris1,Liu3,chen2017video,Wang-CVPR19-Shifting,Wang-PAMI21-Survey,Wang1,Wang2} methods that can better capture image representations. Although these methods have achieved impressive results, they still have computation and memory cost limitations. To address this, FCN-based methods~\cite{9684999,9560713} have been proposed, which formulate SOD as a pixel-wise binary classification task. Recently, several works~\cite{Rynson1,Liu2} have integrated features from multiple layers of CNN to exploit context information at different semantic levels.
In addition, some researchers~\cite{Liu1,Chen-TIP20-Imp,Chen-TIP20-RGBD,Dai1,Dai2} have used depth images as auxiliary information to improve RGB-D SOD performance, but this may be limited in adverse conditions. The development of thermal infrared sensors has facilitated the progress of RGB-T SOD~\cite{Xu-AI22,8695412}, which takes advantage of temperature cues to improve performance in complex scenes. These approaches have attracted much attention and are effective in challenging environments.

\vspace{-0.25cm}
\subsection{Eye Fixation Prediction}
\textbf{E}ye \textbf{f}ixation \textbf{p}rediction (EFP), different from SOD, aims to predict where people get interested in natural scenes. Early EFP models~\cite{itti1998model,cerf2007predicting} used low-level features such as contrast, color, and brightness. However, recent advances in deep neural networks have allowed for learning high-level, top-down features, resulting in significant performance improvements. The SALICON dataset~\cite{jiang2015salicon}, which contains many real human eye observation points, has been used for EFP research~\cite{Wang-TIP17-DVA,Wang-TPAMI19-Inferring}. Generative adversarial networks~\cite{pan2017salgan} and new evaluation metrics have also been introduced into the field. Domain adaptation techniques~\cite{droste2020unified} have been proposed to model image and video tasks in a unified way. However, predicting how objects relate to each other is more challenging than predicting where they are in scenes. This requires achieving two objectives: accurate localization with salient objects and mutual ranking of objects.

\vspace{-0.25cm}
\subsection{Saliency Ranking}
The salient ranking is a newly proposed problem~\cite{Islam-CVPR18-ranking,Rynson1,Liu4} to determine the relative order of different saliency objects. Previous works have made steady progress in this field by adopting various approaches, such as human attention shifts~\cite{Siris-CVPR20-raning}, graph neural reasoning modules~\cite{Liu-TPAMI21-GCN}, and object-context interaction information~\cite{Tian-CVPR22-BI}. Apart from this, Lv et al.~\cite{Lv-CVPR21-camouflaged} propose a new camouflage object detection model to rank camouflaged objects, and they adopt computing the fixation points on the instance to label the instance ranks. Fang et al.~\cite{Fang-ICCV21-PPA} propose an end-to-end SOR framework and introduce a Position-Preserved Attention module that preserves the coordinates of objects in an image.

However, the ground truth of the ranking orders generated by these methods is improper, and the regression-based multi-task learning approach may cause mutual interference between irrelevant tasks. Moreover, implementing regression-based instance segmentation is complex and challenging. 
To address the challenges faced in generating saliency ranking GT orders and conducting regression-based multi-task learning, we propose a comprehensive approach that considers three key areas: GT data generation, network structure design, and training protocol. By addressing these areas, we aim to create a whole-flow processing paradigm for the saliency ranking task.

\begin{figure}[!h]
	\centering
	\includegraphics[width=1.0\linewidth]{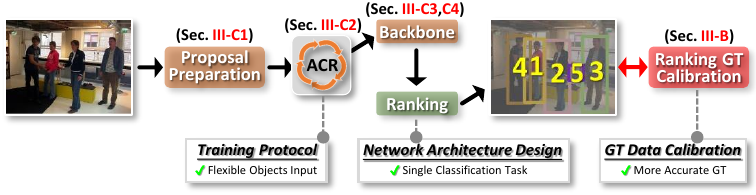}
	\vspace{-0.8cm}
	\caption{Pipeline of our whole-flow processing paradigm of saliency ranking.}
	\vspace{-0.6cm}
	\label{fig:overview}
\end{figure}

\begin{figure*}[!t]
	\centering
	\includegraphics[width=1.0\linewidth]{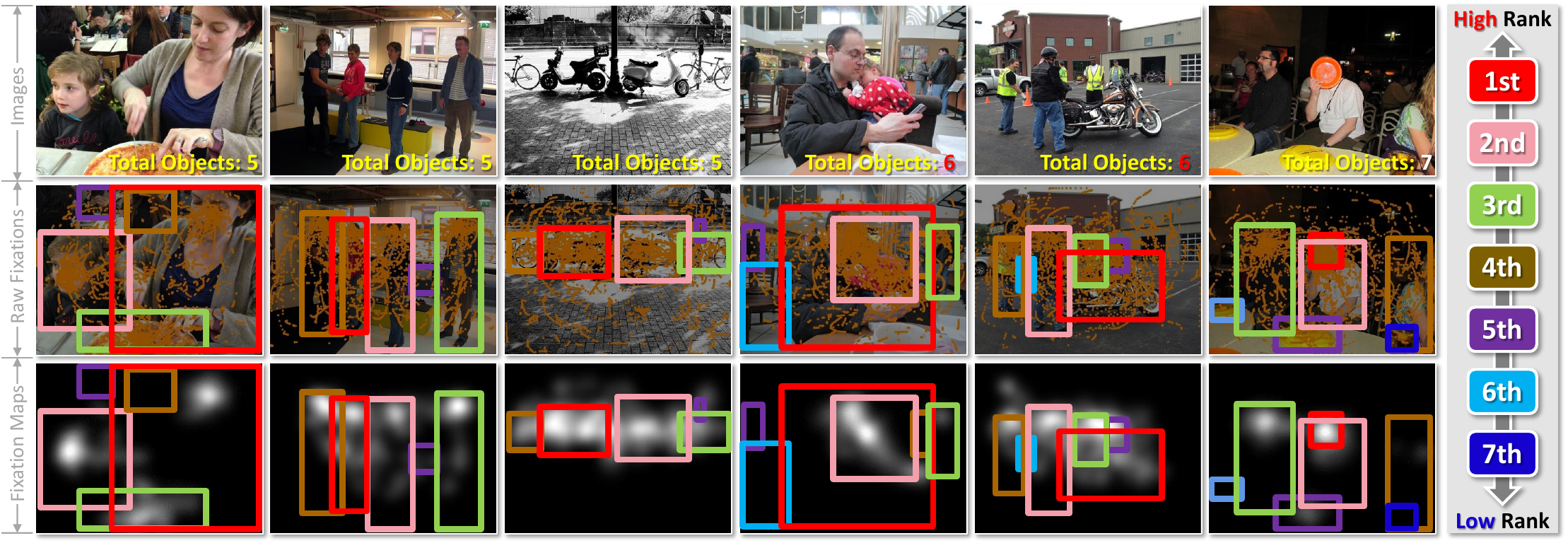}
	\vspace{-0.8cm}
	\caption{Visual comparison between raw fixation points (2nd column) and fixation maps (3rd column), which are commonly used in saliency ranking GT orders generation methods. Saliency ranking orders are marked in rectangular boxes with different colors.}
	\vspace{-0.4cm}
	\label{fig:LGLC}
\end{figure*}

\vspace{0.45cm}
\section{The Proposed Method}


\subsection{Method Overview}
The main objective of our research is to propose a whole-flow processing paradigm of saliency ranking task (refer to Fig.~\ref{fig:overview} for a better understanding) from the perspective of ``\textbf{GT data generation}'', ``\textbf{network structure design}'' and ``\textbf{training protocol}" by addressing the three main issues that arise when generating saliency ranking GT orders (\textbf{Issue1}) and conducting regression-based multi-task learning (\textbf{Issue2} and \textbf{Issue3}). To address \textbf{Issue1}, we propose a \textbf{r}elationship-\textbf{a}ware \textbf{s}aliency \textbf{r}anking \textbf{GT} orders generation method (RA-SRGT, Sec.~\ref{sec:RA-SRGT}) that utilizes varying levels of saliency thresholds to produce saliency ranking GT orders that closely align with the underlying principles of the \textbf{h}uman \textbf{v}isual \textbf{s}ystem (HVS). This is achieved by considering the interrelationships between image regions and incorporating HVS principles into the GT generation process. To address the interference (\textbf{Issue2}) of multi-task learning and complex implementation (\textbf{Issue3}) of regression techniques used in existing saliency ranking methods, we propose a brand-new approach that involves a newly-devised, simple yet efficient proposal-based single-task framework called \textbf{f}lexible \textbf{o}bject \textbf{s}aliency \textbf{r}anking network (FOSRNet, Sec.~\ref{sec:OFSEQ}). Based on classification to better deal with sparser proposals, our approach uses off-line proposals to enhance detection accuracy while reducing computational resources and training complexity compared to existing online proposals generation methods (Sec.~\ref{sec:OD}). Moreover, we introduce an \textbf{a}daptive \textbf{c}irculative \textbf{b}agging (ACB, Sec.~\ref{sec:ACB}) method to solve the problem of flexible object inputs in the classification task. We will provide detailed explanations of each component in the following sections.

\vspace{-0.3cm}
\subsection{Relationship-aware Saliency Ranking GT Order Generation}
\label{sec:RA-SRGT}

Salient object detection and eye fixation prediction can produce saliency maps of multiple objects. Still, these maps are limited in their ability to express these objects' relationships and degrees of saliency. Since these methods can only locate salient areas and do not distinguish which is more salient among them, consequently, some researchers use saliency ranking to address this challenge. However, the saliency ranking GT orders are assigned in various ways, and erroneous GT orders can negatively impact model training and fail to align with human attention and psychology studies (\textbf{Issue1}).
To further illustrate the impact of different saliency ranking GT orders generation methods, we shall briefly detail two commonly-used methods: fixation points-based and fixation maps-based. Note that shift attention approaches are excluded because they differ from relative saliency.

\begin{figure*}[!t]
	\centering
	\includegraphics[width=1.0\linewidth]{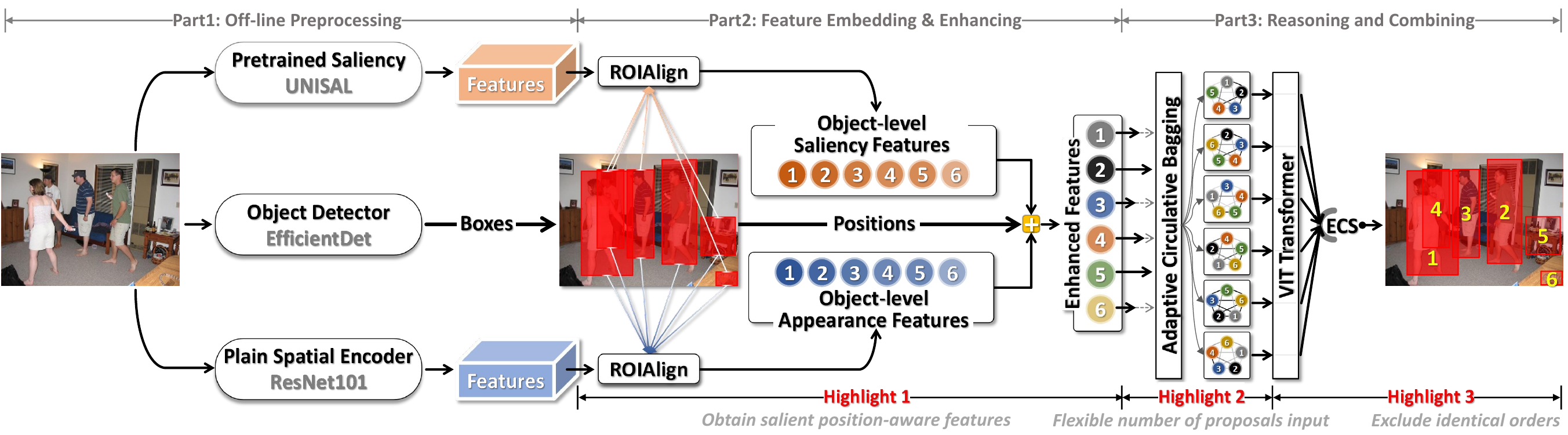}
	\vspace{-0.7cm}
	\caption{The pipeline of our flexible object saliency ranking network. The ranking model mainly consists of three parts: Part1) off-line preprocessing to prepare off-line proposals and features for later processing; Part2) feature embedding \& enhancing to embed and improve the object-level appearance and saliency features; Part3) reasoning and combining to settle down the flexible multiple objects input by the \textbf{a}daptive \textbf{c}irculative \textbf{b}agging (ACB) and implement classification with an \textbf{e}xclusive \textbf{c}la\textbf{s}sification module (ECS) to avoid the same ranking order existing in traditional classification tasks. ``$\oplus$'' is the feature concatenation operation.}
	\vspace{-0.5cm}
	\label{fig:pipeline}
\end{figure*}

1) \textbf{Fixation points-based saliency ranking GT orders generation approaches} (Fig.~\ref{fig:LGLC}-line 2). These methods mainly rely on counting the total number of fixation points within each object and assigning higher saliency ranking orders to objects with more fixation points, \emph{i.e.}, the saliency ranking orders of objects in an image are determined by the number of fixation points and the spatial size of the objects. This can be formulated as:
\begin{equation}
	{\rm Rank}({\rm Q}_i) = \underset{\rm Penalty} {\underbrace{\frac{1}{\sqrt{{\rm size}({\rm Q}_i)}}}} \overset{\rm Total\ Fixations\ in\ {\rm Q}_{\emph i} }{\overbrace{\sum_{(u,v)\in {\rm Q}_i} {\rm P}(u,v)}},\\[-0.8ex]
\end{equation}
where ${\rm Q}_i$ denotes the $i$-th object; ${\rm Rank}({\rm Q}_i)\in\{1,2,...,n\}$ ($n$ is the total number of objects in an image scene) returns the saliency ranking order of ${\rm Q}_i$; ${\rm Size}({\rm Q}_i)$ measures the spatial size of ${\rm Q}_i$, \emph{i.e.}, width$\times$height; ${\rm P}(u,v)\in\{0,1\}$ indicates whether there exists a fixation point at spatial coordinate $(u,v)$. However, these fixation points-based approaches only consider intra-object relationships, which means they focus on isolated objects without considering inter-object relationships, \emph{i.e.}, how those objects relate to one another and the overall composition of the image, leading to inaccurate saliency ranking orders (Mark {\large{\textcircled{\small{1}}}} in Fig.~\ref{fig:motivation1}-A).

2) \textbf{Fixation maps-based saliency ranking GT orders generation approaches} (Fig.~\ref{fig:LGLC}-line 3). These methods assign saliency ranking GT orders by calculating the maximum/average pixel values of each object in fixation maps and assign higher saliency ranking orders to objects with larger pixel values. This process can be denoted as follows:
\begin{equation}
	{\rm Rank}({\rm Q}_i) =  \underset{\rm Maximum\ pixel\ values}{\underbrace{{\rm MAX}\big({\rm V}(u,v)\big)}}\ {\rm or}\  \underset{\rm Average\ pixel\ values}{\underbrace{{\rm AVG}\big({\rm V}(u,v)\big)}},
\end{equation} 	
where ${\rm V}(u,v)\in\{0,255\}$ denotes the pixel value at spatial coordinate $(u,v)$ with the $i$-th object ${\rm Q}_i$ in a fixation map. $\rm MAX/AVG$ is to calculate the maximum/average pixel value of ${\rm Q}_i$.
Whereas, as shown in Fig.~\ref{fig:motivation1}-A-Mark {\large{\textcircled{\small{2}}}}, the biggest problem of fixation maps-based methods is the inconsistent ranking results across different saliency ranking GT orders generation methods and sensitivity to noise occurring in fixation maps.

As mentioned in Fig.~\ref{fig:motivation1}-A, the limitations of fixation points-based and fixation maps-based saliency ranking GT orders generation methods result in unreliable and inconsistent saliency ranking GT orders unaligned with the principles of the human visual attention system. These methods overlook the overall image composition and inter-object relationships, crucial in precisely reflecting human attention and psychology.

3) \textbf{Relationship-aware saliency ranking GT orders generation}.
To address the limitation mentioned above, from the perspective of ``\textbf{GT data generation}'', we propose a novel method called \textbf{r}elationship-\textbf{a}ware \textbf{s}aliency \textbf{r}anking \textbf{GT} orders generation (RA-SRGT). This method is fixation points-based and considers intra-object and inter-object relationships among objects, enabling a comprehensive evaluation of saliency ranking. By considering the entire image composition, RA-SRGT generates reliable and consistent saliency ranking orders that accurately reflect human attention and psychology.

Intuitively, there are two ways to implement our RA-SRGT. The first way involves binarizing the fixation maps after graying and counting the white pixels of each object. The percentage of each object in the entire image is then calculated, and the final saliency ranking order is obtained by multiplying these two parts. The whole process can be defined as:
\vspace{-0.1cm}
\begin{equation}
	{\rm Rank}({\rm Q}_i) = {\rm Sort}\bigg(\sum_{i=1}^{n} \frac{{\rm P}_{{\rm Q}_{i}}^{white}}{{\rm P}_{{\rm Q}_{i}}^{total}} \times \frac{\sqrt{{\rm size}({\rm Q}_i)}}{\sqrt{{\rm size}({\rm Q})}} \bigg),
\end{equation}
where ${\rm Q}_i$ denotes the $i$-th object of the image scene ${\rm Q}$; ${\rm Rank}({\rm Q}_i)\in\{1,2,...,n\}$ ($n$ is the total number of objects in an image scene) is the final saliency ranking GT order of ${\rm Q}_i$; ${\rm Sort}(\cdot)$ represents the sorting operation, returning the saliency ranking order of ${\rm Q}_i$; ${\rm Size}(\cdot)$ measures the spatial size; ${\rm P}_{{\rm Q}_{i}}^{white}$ and ${\rm P}_{{\rm Q}_{i}}^{total}$ separately indicate the number of white pixels and all pixels within the binary fixation map of $\rm Q$.

Compared to existing fixation maps-based saliency ranking GT orders generation approaches, this binarized fixation maps-based scheme considers both intra-object and inter-object relationships among objects. But it still has a limitation, as it is sensitive to noise. The binary saliency maps with noise may not accurately reflect each object's true saliency, leading to ranking errors. Moreover, the binary threshold may vary depending on the saliency maps' quality and content, making it hard to generalize.

Therefore, we adopt the second option, which utilizes GT saliency annotations, \emph{i.e.}, fixation points generated by mouse trajectories provided in the SALICON set~\cite{SALICON}. These fixation points-based saliency annotations are used for eye fixation prediction tasks, and we resort to them to work for our saliency ranking. The main superiority of the fixation points-based way to implement RA-SRGT compared to the above-mentioned binarized fixation maps-based way is that counting fixation points are more reliable and controllable than binarizing fixation maps, which requires selecting suitable binary thresholds.
Moreover, the key difference between our fixation points-based RA-SRGT and the existing fixation points-based saliency ranking GT orders generation methods mentioned above is that we do not rely solely on the number of fixation points. Instead, we consider the percentage of each object in the entire image.

Unlike the first scheme to count white pixels within binarized fixation maps, we count the fixation points of each single rectangle proposal, which object detection models generate. The motivation behind this is to provide a more accurate and reliable way to rank the saliency of objects in an image.
By taking into account the relative importance of each object in the image, this approach is less sensitive to noise and threshold values, which can vary depending on the image's content and quality. Additionally, this approach reflects that the human visual system pays attention to objects that are contextually important rather than simply visually salient.
In summary, our approach follows the process outlined below:
\begin{equation}
	\begin{split}
{\rm {\small Total\ Fixations\ in\ {\rm Q}_{\emph i}}}\ \ \ \ \ \ \ \ \ \ \ \ \ \ \ \ \ \ \ \ \ \ \ \ \ \ \ \ \ \ \ \ \ \ \ \\[-0.8ex]
		{\rm S}({\rm Q}_i)  = \begin{cases}
{\overbrace{\sum_{(u,v)\in {\rm Q}_i}{\rm P}(u,v)}} + \underset{\rm Penalty}{\underbrace{\gamma  \cdot {\rm e}^{\beta \cdot \sqrt{{\rm size}({\rm Q}_i)}}}},
			&\text{ if } {\rm N}_{i}>0, \\
			\ \ \ \ \ 0 ,    &\text{ if } {\rm N}_{i}  = 0,
		\end{cases}
	\end{split}
	\label{eq:gammabeta}
\end{equation}
where ${\rm S}({\rm Q}_i)$ returns the combined score of the $i$-th object ${\rm Q}_i$; ${\rm P}(u,v)\in\{0,1\}$ indicates whether there exists a fixation point at spatial coordinate $(u,v)$ of ${\rm Q}_i$; $\sqrt{{\rm size}({\rm Q}_i)}$ means the spatial size of ${\rm Q}_i$.
To mitigate the influence of various factors on the ranking scores, we propose a set of thresholds $\gamma$ and $\beta$. The purpose of $\gamma$ is to serve as the GT threshold (ablation study can be found in Fig.~\ref{fig:userstudy}-B), while $\beta$ is to align ${\rm P}(u,v)$ numerically, a value that can be considered as being hidden within $\gamma$ (ablation study can be found in Fig.~\ref{fig:userstudy}-C). Specifically, $\gamma$ controls the extent to which the spatial size of the object affects the ranking, while $\beta$ determines the impact of the fixation points. By varying the values of $\gamma$ and $\beta$, we can analyze how these factors interact and identify the optimal combination for generating the saliency ranking GT orders.

Since we use object proposals instead of instance segmentation maps to represent objects (see Sec.~\ref{sec:OD} for a reason), some proposals may have no fixation points. Therefore, if the fixation number of a proposal is zero, we label all such proposals as novel ranking orders, \emph{i.e.}, 0.
Then we can get the final saliency ranking GT orders ${\rm Rank}({\rm Q}_i)$ of ${\rm Q}_i$ by applying:
\begin{equation}
	\begin{split}
		{\rm Rank}({\rm Q}_i) = \begin{cases}
			\underset{i\in (1,n)}{\underbrace{{\rm Sort}\big({\rm S}({\rm Q}_i)\big)}}, &\text   {if}\ {\rm S}({\rm Q}_i)>0 \\
			0,  &\text   {if}\ {\rm S}({\rm Q}_i)=0
		\end{cases}
	\end{split}
\end{equation}
where ${\rm Sort}(\cdot)$ represents the sorting operation, returning the saliency ranking order of ${\rm Q}_i$ and $n$ is the total number of objects in an image scene.

\textbf{Summary}.
Our proposed methods offer a more precise solution for obtaining accurate saliency ranking GT orders than fixation maps, which rely on arbitrary binary thresholds. By using GT saliency annotations in the form of fixation points, we can avoid such issues of sensitivity of noise, as each observer's annotation data is independent and exact. This approach enables us to obtain highly reliable identification of salient regions in an image across multiple observers. Therefore, our newly-proposed methods significantly improve over previous approaches and can provide valuable insights for various applications in computer vision and beyond. 

\vspace{-0.25cm}
\subsection{Flexible Object Saliency Ranking Network}
\label{sec:OFSEQ}

Multi-task learning refers to training a single model to perform multiple tasks simultaneously. In this approach, the model learns to share the learned representation across different tasks, which can improve the model's overall performance. However, a significant challenge in multi-task learning is that the different tasks may interfere, leading to unstable training and suboptimal performance (\textbf{Issue2}).
For instance, when combining segmentation/object detection and saliency ranking, the segmentation/detection task may affect the saliency ranking task, making the model more complex and difficult to train. Moreover, current saliency ranking methods rely heavily on regression-based instance segmentation, a pixel-wise approach that demands a large amount of data to yield precise outcomes. This technique can be intricate and challenging to implement effectively, posing a significant challenge (\textbf{Issue3}).

To address these issues, we propose a single-task-based framework called \textbf{f}lexible \textbf{o}bject \textbf{s}aliency \textbf{r}anking network (FOSRNet) from the perspective of ``\textbf{network structure design}''. This framework simplifies the multi-task learning process by training the model to simultaneously perform a single saliency ranking task. It also solves the complex of regression-based methods and saliency ranking tasks by a classification-based architecture.

As visualized in Fig.~\ref{fig:pipeline}, FOSRNet consists of three main parts: off-line preprocessing (\textbf{Part1}), feature embedding \& enhancing (\textbf{Part2}), and reasoning and combining (\textbf{Part3}). In \textbf{Part1}, we generate object proposals offline. We avoid complicated training while retaining high detection accuracy by leveraging the already high-performance and pre-trained object detection models to obtain object proposals offline. In \textbf{Part2}, we utilize a pre-trained saliency model (\emph{e.g.}, UNISAL) and plain spatial encoder (\emph{e.g.}, ResNet101) to provide object-level saliency features and appearance features, respectively, and jointly embed them to object-level enhanced salient position-aware features for latter classification.
Finally, in \textbf{Part3}, we address the issue of complex implementation of regression-based ranking models. To do so, we employ \textbf{e}xclusive \textbf{c}la\textbf{s}sification (ECS) as described in Section \ref{sec:ECS}. ECS ensures that different objects are predicted with unique saliency ranking orders, which is necessary for a classification task. This approach helps to improve the accuracy and reliability of the classification results. Particularly, we use \textbf{a}daptive \textbf{c}irculative \textbf{b}agging (ACB, Sec.~\ref{sec:ACB}) to process the flexible number of input proposals circulatively, which is more applicable to real-world scenarios where the input is dynamic and complex. We will elaborate on each part in the following.

\subsubsection{Off-line Preprocessing}
\label{sec:OD}
\begin{figure}[!t]
	\centering
	\includegraphics[width=1.0\linewidth]{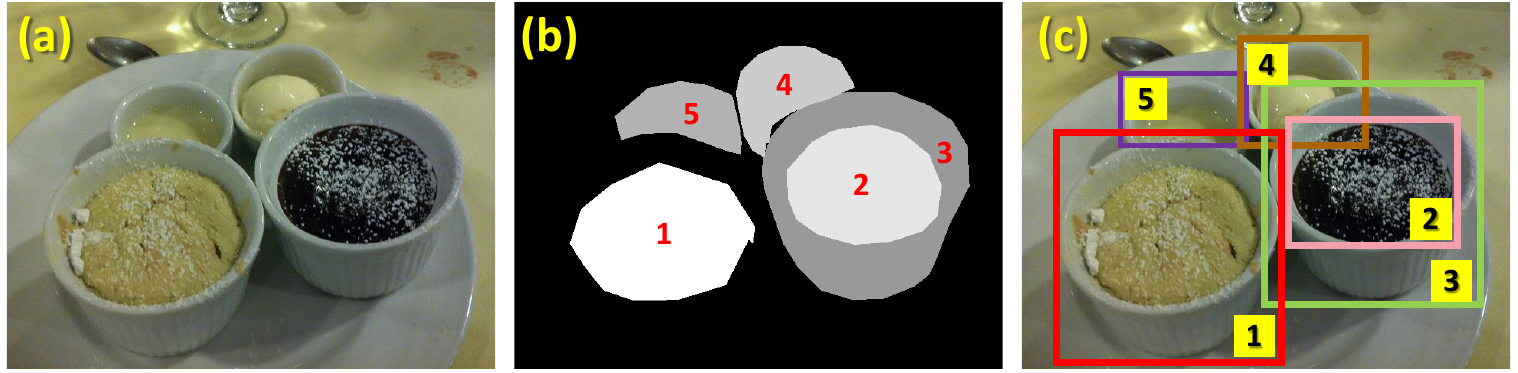}
	\vspace{-0.7cm}
	\caption{Comparison of the saliency ranking orders between instance segmentation-based (b) and object detection-based saliency ranking (c).}
	\label{fig:SegDetComp}
	\vspace{-0.4cm}
\end{figure}

Most existing saliency ranking methods are based on instance segmentation or object detection tasks, as shown in Fig.~\ref{fig:SegDetComp}. Among them, instance segmentation tasks require producing pixel-wise masks that can be challenging. The ranking order is highly sensitive to the quality of the instance masks, and inaccurate ranking orders can result from poor-quality masks.
And object detection tasks tend to perform online proposal generation, which increases the complexity of model training and leads to inferior saliency ranking results.

Therefore, instead of relying on pixel-wise instance segmentation and heavy online object detection, we focus on off-the-shelf object-wise proposals~(Fig.~\ref{fig:pipeline}-\textbf{Part1}), which is a simpler task that can be effectively performed offline. To achieve off-line generation, we use an off-the-shelf state-of-the-art (SOTA) object detection method, such as EfficientDet\footnote{Other SOTA object detection models can replace it, and the ablation study is shown in Table.~\ref{Robustness}.}~\cite{tan2020efficientdet} to decompose the given image into at least five rectangular object proposals\footnote{We assume that an image has at least five object proposals since most images (nearly 70\%) in SALICON set contain about five objects. When the proposed model processes an image containing fewer than five objects, it automatically fills to five objects using empty objects, a.k.a. dummy nodes.}.
Then, filtering out overlapping object proposals with low \textbf{i}ntersection \textbf{o}ver \textbf{u}nion (IOU) rate between each pair of proposals. Additionally, we discard extremely large object proposals (those with an object proposal area proportion greater than 60\% of the whole image area) and extremely small object proposals (those with a sum of pixel values less than 20). All these processes are conducted offline, avoiding complicated training while retaining high detection accuracy.
We also resort to a pre-trained saliency model and plain spatial encoder to obtain enhanced salient position-aware features for the latter classification, which will be detailed later.

\subsubsection{Feature Embedding \& Enhancing}
\label{sec:SPAFE}

After obtaining proposals from Sec.~\ref{sec:OD}, several feature backbones are available, \emph{e.g.}, CNN, RNN, Transformer, GNN, and MLP. Here we select Transformer (\emph{e.g.}, VIT~\cite{VIT}) as our feature extractor due to its powerful modeling of long-range dependency, which is significant to explore the inter-object relationships in a scene. Unlike traditional plain patch embedding methods such as image-level partition or CNN-based feature-level partition in Transformer, we propose to infuse salient position-aware knowledge inherited from fixations and deep appearance features from deep feature extractors to the embedded features (Fig.~\ref{fig:pipeline}-\textbf{Part2}). This knowledge is position-aware and saliency-aware, meaning it can establish the relationships between the individual proposal and the whole image, which can help accurately rank the objects.

\begin{figure}[!t]
	\centering
	\includegraphics[width=1.0\linewidth]{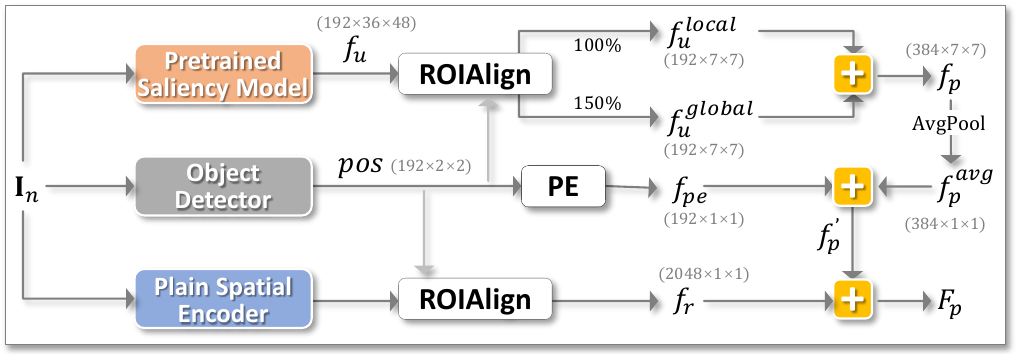}
	\vspace{-0.8cm}
	\caption{Feature embed \& enhance. PE: Position Embed; $\oplus$: Concatenate.}
	\label{fig:SPAFE}
	\vspace{-0.6cm}
\end{figure}

The feature embedding and enhancing process shown in Fig.~\ref{fig:SPAFE} comprises two parts: a \textbf{p}retrained \textbf{s}aliency \textbf{m}odel (PSM) and a \textbf{p}lain \textbf{s}patial \textbf{e}ncoder (PSE).
The PSM extracts object-level saliency features that inherit salient position-aware knowledge from fixations. This is achieved using the off-line position-aware eye fixation prediction model UNISAL~\cite{droste2020unified}. On the other hand, the PSE obtains object-level appearance features that contain deep spatial and semantic information. This is done using the off-the-shelf pre-trained ResNet101 to obtain high-dimensional semantic deep features. The object-level saliency features and appearance features work together to enhance the feature embedding process for improved performance.
We will explain them below.

The PSM takes the whole original image $I_{n}$ as input, outputting  $f_{u}$, which then is fed into ROIAlign to produce two proposals of different scales: a local feature-level proposal $f_{u}^{local}$ (same size as $f_{u}$), and a global feature-level proposal $f_{u}^{global}$ (50\% larger than $f_{u}$). The global feature-level proposal can capture more global information from the original image and enhance the proposal features. The ROIAlign operation requires exact proposal location information, which is lost in $f_{u}$. We have added extra location information (denoted as $pos\in \mathbb{R}^{192\times 2\times 2}$, containing coordinates of the top-left and bottom-right corners of each proposal) in it. Then, fuse the two feature-level proposals to obtain salient position prior-oriented feature $f_{p}$\footnote{$f_{p}$, generated by the output of PSM ($f_{u}$), naturally possesses saliency information, since PSM is derived from EFP method, which is also to predict salient regions.}.
This process can be formulated as follows:
\begin{equation}
	\begin{aligned}
		&\ \ \ \ \ \ \ \ \ \ \ \ \underset {\ \Downarrow\ }{\underbrace{{\rm ROIAlign\big(}(f_{u}), 1.5r , pos\big)}} \\[-0.5ex]
		&f_{p} ={\rm Concat}(f_{u}^{local},f_{u}^{global}),\ f_{u}={\rm PSM}(I_{n}),\\[-0.8ex]
		&\ \ \ \ \overbrace{{\rm ROIAlign\big(}(f_{u}),\ r, \ pos\big)}^{\Uparrow}	\\[-0.5ex]
	\end{aligned}	
\end{equation}
where, $r$ is the region of interest (ROI) from image-level object proposals; $\rm Concat(\cdot)$ denotes feature concatenation; $\rm ROIAlign(\cdot)$ is to extract ROIs and map them to a fixed size.

\begin{figure*}[!t]
	\centering
	\includegraphics[width=\textwidth]{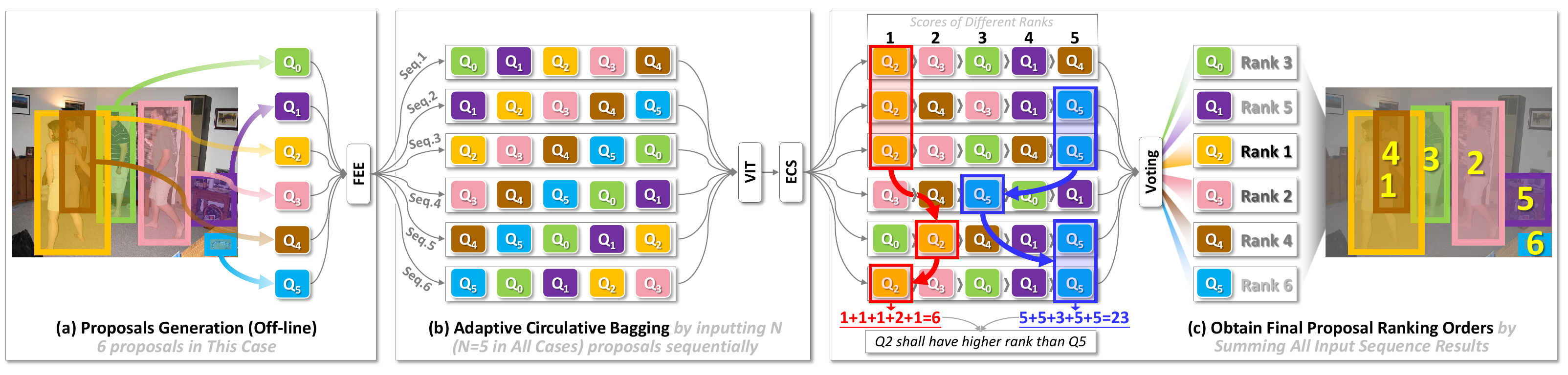}
	\vspace{-0.9cm}
	\caption{Pipeline of adaptive circulative bagging. The process includes two phases: adaptive circulative bagging (b) and final proposal ranking orders obtaining (c). Specifically, the times of circulative bagging are determined by the number of proposals, \emph{e.g.}, six times in the case shown in the figure. And each time, we input five proposals in all cases. The final ranks are achieved by summing the ranking score results of each proposal's six times circulative input and voting each proposal's final saliency ranking order according to the ranking scores (smaller scores, smaller orders, higher saliency ranking/importance).}
	\label{fig:ACB}
	\vspace{-0.4cm}
\end{figure*}

Note that salient position prior-oriented feature $f_{p}$ is inherited from $f_{u}$ generated by PSM, \emph{i.e.}, UNISAL, it naturally contains a small amount of location information. However, the location information is not accurate. To enhance the position information among proposals, we add each proposal's central point coordinate information to the feature, which begins by scaling down the feature-level proposal $f_{p} \in \mathbb{R}^{384\times 36\times 48} $ to the size of $f_{p}^{avg}\in \mathbb{R}^{384\times 1\times 1} $ through global average pooling. Then, integrate it with position embedding $f_{pe}$, which represents the center coordinate (x,y) of $f_{p}^{avg}$. The object-level saliency feature $f_{p}^{'}\in  \mathbb{R}^{576\times 1\times 1}$ can be formulated as:
\begin{equation}
	\begin{aligned}
		&f_{p}^{'} ={\rm Concat}(f_{p}^{avg},\ f_{pe}),\ f_{p}^{avg}= {\rm AvgPool}(f_{p})\\
	\end{aligned}	
\end{equation}
where ${\rm AvgPool}(\cdot)$ is the average pooling operation; $f_{pe}=\rm PE(\cdot)$ denotes position embedding consisting of a $1\times1$ convolution to make position information $pos$ learnable.

However, though object-level saliency feature $f_{p}^{'}$ is equipped with enhanced position information, it only contains shallow features without deep semantic information since the PSM is a lightweight architecture that can merely generate shallow position-related features.
To overcome, we resort to a \textbf{p}lain \textbf{s}patial \textbf{e}ncoder (PSE) to obtain object-level appearance feature $f_{r} \in \mathbb{R}^{2048\times 1\times 1}$ that contain deep spatial and semantic information to compensate for the lack of high-level semantic cues.
Finally, a more powerful object-level enhanced salient position-aware feature $F_{p}\in \mathbb{R}^{2642\times 1\times 1}$, which contains object-level saliency features and deep object-level appearance features simultaneously is generated by fusing $f_{p}^{'}$ and $f_{r}$. It can be formulated as follows:
\begin{equation}
	F_{p} ={\rm Concat}(f_{p}^{'},\ f_{r}),\\
\end{equation}

Then, the object-level enhanced salient position-aware feature $F_{p}$ will be used to predict saliency ranking orders. Next, we will detail how to use this enhanced feature to implement a saliency ranking task.

\subsubsection{Reasoning and Combining}
\label{sec:RC}
As detailed in Sec.~\ref{sec:SPAFE}, we select Transformer (\emph{e.g.}, VIT~\cite{VIT}) as our feature extractor due to its powerful modeling of long-range dependency, which is significant to explore the inter-object relationships in a scene. However, there exists a flexible input proposals issue, which conventional Transformer networks cannot solve. Thus, adapting it to a classification-based saliency ranking method with a fixed number of proposal classes, \emph{i.e.}, a saliency ranking network with a fixed input, remains challenging. This motivates us to conduct further exploration, as seen below.

\vspace{0.1cm}
\noindent\textbf{Adaptive Circulative Bagging}.
\label{sec:ACB}
Regression-based saliency ranking methods are a more practical choice in real-world scenarios because they can handle varying numbers of objects in an image. Unlike classification-based methods, which require a fixed number of proposal classes, regression models predict the saliency scores of each object directly based on its visual features. This means they can analyze any number of objects in the image without needing a predefined number. In contrast, classification-based methods may be limited in recognizing a fixed number (\emph{e.g.}, 5) of objects and may not predict saliency ranking orders for scenes with more than or less than the fixed number of objects.
To address the limitation of classification-based methods, concerning ``\textbf{training protocol}", we propose a brand-new method called \textbf{a}daptive \textbf{c}irculative \textbf{b}agging (ACB, Fig.~\ref{fig:pipeline}-\textbf{Part3}) to handle varying numbers of input proposals from any object detection model in a circulative manner.

The ACB each time processes five object-level enhanced salient position-aware features (each corresponds to a proposal within an image) obtained by Feature Embedding \& Enhancing~(Fig.~\ref{fig:pipeline}-\textbf{Part2}). Suppose there are more than five object-level features. In that case, we start by selecting the object-level features ${\rm Q}_{0}$\footnote{Here, we start the subscript from 0 instead of 1 to fit better the following formula (eq.~\ref{eq:acb}).} as the center object-level features and sort the other object-level features based on their distance to ${\rm Q}_{0}$ in ascending order. Then, we process each object-level feature in sequence, separately starting from ${\rm Q}_{1}$, ${\rm Q}_{2}$, and so on, until all object-level features have been processed.
To better understand the process, please refer to Fig.~\ref{fig:ACB}-(b).
The whole processing phase is as follows:
\begin{equation}
	{\rm Seq}_{i} = \textstyle \bigcup_{k=0}^{4} {\rm Q}_{\{(i+k) \bmod n\}}, \quad 0\le i\le n-1
	\label{eq:acb}
\end{equation}
where ${\rm Seq}_{i}$ refers to the $i$-th input object-level features list sequence of an image; $\textstyle \bigcup$ denotes the union operation of object-level feature sequences by order, and $\{(j+k) \bmod i\}$ means circularly taking the remainder. $n$ is the total number of proposals in an image.
Before feeding the object-level features list sequence ${\rm Seq}_{i}$ into the classification reasoning part, there exists a fatal issue --- same classification results (\emph{i.e.}, same saliency ranking orders for objects with different saliency ranking orders in traditional classifications to be solved.

\vspace{0.1cm}
\noindent\textbf{Exclusive Classification}.
\label{sec:ECS}
Current saliency ranking methods rely heavily on complex regression-based instance segmentation, a pixel-wise approach requiring significant data to produce accurate results. Consequently, this technique can be challenging, leading to a substantial obstacle (\textbf{Issue3}). Thus, we choose a classification to develop effective ranking models.

Since saliency ranking is similar to multiple classification tasks, intuitively, we can use two \textbf{f}ully-\textbf{c}onnected layers (FC) to generate the final saliency ranking orders directly. However, traditional classification can suffer from a fatal issue, \emph{i.e.}, due to the limitation of the ``Max'' function used in FC layers to predict classes, it might assign the same classification results (\emph{i.e.}, same saliency ranking orders) for objects with different saliency ranking orders. For example, in a scene with six objects, which should be assigned six distinct saliency ranking orders, inversely, traditional classification tasks might assign two objects with the same order, which does not fit the essence of the saliency ranking task.

Therefore, in order to achieve a more accurate and reliable saliency ranking, we employ an \textbf{e}xclusive \textbf{c}la\textbf{s}sification (ECS) method to predict the saliency ranking orders of proposals in each iteration. The ECS includes two FC layers, a SoftMax function, and a Hungarian algorithm to handle the same ranking order issue. Since the Hungarian algorithm is used to solve the bipartite graph matching problem, which can ensure that each category contains unique data, thus, solving the problem of identical saliency ranking orders in conventional classification problems.

Each proposal's final saliency ranking order can be collected by: 1) summing the saliency ranking score outputs of the total number of $n$ input object-level features list sequences and 2) voting each proposal's final saliency ranking order (smaller scores, smaller orders, higher saliency ranking/importance). As shown in Fig.~\ref{fig:ACB}-(c), the whole processing can be denoted as:
\vspace{-0.8cm}
\begin{equation}
	\begin{aligned}
		&{\rm Rank}({\rm Q}_{j}) = {\rm Sort}\bigg( {\textstyle \sum_{j=0}^{n-1}} {\textstyle \sum_{i=0}^{n-1}}\Big({\rm Net}\big({\rm Seq}_{i,j}\big)\Big)\bigg),\\[-2.0ex]
		&\ \ \ \ \ \ \ \ \ \ \ \ \ \ \ \ \ \ \ \ \ \ \ \ \ \ \ \overbrace{{\rm ECS}\bigg({\rm VIT}\Big({\rm ACB}\big({\rm FEE}({\rm Seq}_{i,j})\big)\Big)\bigg)}^{\Uparrow}	\\[-2.0ex]
		&\ \ \ \ \ \ \ \ \ \ \ \ \overbrace{{\rm Hungary\Big(SoftMax\big(FC}({\rm F}_{g})\big)\Big)}^{\Uparrow}
		\label{eq:HA}
	\end{aligned}	
\end{equation}
where ${\rm Rank}({\rm Q}_{j})$ is the final saliency ranking order of the $j$-th object-level feature (\emph{i.e.}, proposal) ${\rm Q}_{j}$; ${\rm Seq}_{i,j}$ means the $i$-th input proposal list sequence of $j$-th proposal of an image, where $i!=(j-1+n)\ mod\  n$; ${\rm F}_{g}$ means the output of $\rm VIT(\cdot)$; $\rm FEE(\cdot)$ is the feature embedding \& enhancing~(Sec.~\ref{sec:SPAFE}); $\rm ACB(\cdot)$ is the adaptive circulative bagging~(Sec.~\ref{sec:ACB}); ${\rm Sort}(\cdot)$ represents the sorting operation, returning the saliency ranking order of ${\rm Q}_{j}$.

\begin{figure*}[!t]
	\centering
	\includegraphics[width=1\linewidth]{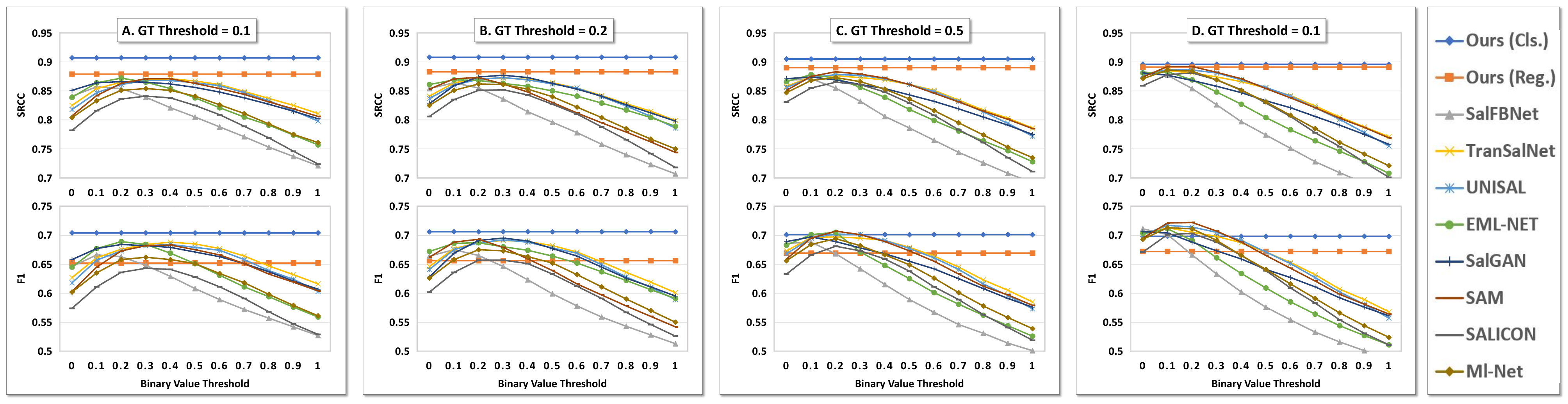}
	\vspace{-0.8cm}
	\caption{Performance illustration of the proposed method with other SOTA models in terms of SRCC (upper line) and F1 (lower line) curves with the binary value threshold of saliency maps at four GT thresholds ($\gamma$ = 0.1, 0.2, 0.5 and 0.8 from left to right).}
	\vspace{-0.4cm}
	\label{fig:SROCC}
\end{figure*}

\subsubsection{Loss Function}
The essence of our saliency ranking network is a typical classification task; thus, we use CrossEntropy Loss (${\rm L}_{cls}$) and MarginRanking Loss (${\rm L}_{rank}$) to optimize the model. The total loss (${\rm L}_{all}$) is defined as follows:
\begin{equation}
	\begin{aligned}
		&\ \ \underset{ \Downarrow}{\underbrace{-\frac{1}{N} {\textstyle \sum_{i=0}^{N-1}} {\textstyle \sum_{j=0}^{N-1}} t_{i,j} {\rm log}(\hat{y}_{i,j}) }}\\[-1.0ex]
		&\ \ \ \ \ \ \ \ \	{\rm L}_{all} = {\rm L}_{cls} + \alpha \cdot{\rm L}_{rank}(\hat{y}_{j},\hat{y}_{i},z),\\[-1.0ex]
		&\ \ \ \ \ \ \ \ \ \ \ \hspace{0.1cm}\ \overbrace{\textstyle \sum_{i,j}^{N}{\rm max}(0,-z\times (\hat{y}_{j}-\hat{y}_{i})+m)}^{\Uparrow }\\[-0.5ex]
	\end{aligned}
	\label{eq:tigd}
\end{equation}
where $\hat{y}_{j}$ and $\hat{y}_{i}$ are the predicted outputs, while $z$ is the true label belonging to either $\left \{ 1 , -1 \right \}$. ``1'' indicates that $\hat{y}_{j}$ has a higher rank than $\hat{y}_{i}$, and ``-1'' means that $\hat{y}_{j}$ has a lower rank than $\hat{y}_{i}$. The minimum value of the correct ranking difference, denoted by $m$, is set to ``0'' in this case.


\vspace{-0.25cm}
\section{Experiments}

\subsection{Datasets and Evaluation Metrics}
Following the approach in~\cite{Siris-CVPR20-raning}, we have relabeled the SALICON dataset using our proposed RA-SRGT~(Sec.~\ref{sec:RA-SRGT}). Specifically, we divided the SALICON validation set into a validation set and a test set since the SALICON test set did not contain fixation points that our RA-SRGT method could utilize. The resulting training, validation, and test sets contained 10,000, 1,200, and 3,800 samples. We adopt the \textbf{s}pearman \textbf{r}ank-order \textbf{c}orrelation \textbf{c}oefficient (SRCC) and the F1 score to evaluate the performance of our method. The higher the SRCC and F1 score, the better the ranking performance.

\vspace{-0.25cm}
\subsection{Implementation Details}
We implement our approach in Python with the Pytorch toolbox on an NVIDIA GTX2080Ti GPU (with 11G RAM). We optimize the network via SGD with a momentum of 0.9 and weight decay of $10^{-4}$. The learning rate is set to 0.001 and exponentially decayed by 0.1 after each ten epoch. The batch size is set to 5 since each image has at least five objects. The weights of UNISAL and ResNet101 are frozen during the whole training phase. The number of Transformer encoder blocks is set as 4 in our network.


\vspace{-0.25cm}
\subsection{Comparison with State-of-the-Art Methods}

\subsubsection{Comparison with Eye Fixation Prediction Methods}
\label{CompEFP}
While diverse objects and instances are generated by object detection or segmentation methods among saliency ranking methods, it can be challenging to make a direct comparison. Therefore, we conducted a quantitative comparison mainly with SOTA eye fixation prediction models. To demonstrate the effectiveness of the proposed method, we have compared it with the eight most recent SOTA eye fixation prediction models, \emph{e.g.}, SalFBNet$_{22}$~\cite{SalFBNet}, TranSalNet$_{21}$~\cite{TranSalNet}, UNISAL$_{20}$~\cite{droste2020unified}, EML-NET$_{20}$~\cite{EML-NET}, SAM$_{18}$~\cite{SAM}, SalGAN$_{18}$~\cite{SalGAN}, ML-Net$_{16}$~\cite{ML-Net}, and SALICON$_{16}$~\cite{SALICON}. To ensure a fair comparison, all quantitative evaluations were conducted using the saliency maps provided by the authors or obtained from the available codes, with parameters unchanged and trained on the SALICON dataset.



Eye fixation prediction methods only generate saliency maps. Therefore, we followed previous saliency ranking methods to obtain saliency ranking orders of other methods by counting the number of white pixels within each object after graying and binarizing the fixation maps. The more white pixels within an object, the more salient the object is. However, the binary threshold obtained by averaging all the values of the grayed whole image in other methods could lead to a false alarm issue, where an object is not salient, but the number of white pixels might still be relatively high.

We propose obtaining a more accurate binary threshold to generate binary maps to ensure a fair comparison. First, we gray (denoted as $\Theta$) the saliency map. Next, we separately calculated the average values of each object in an image by summing the grayed values of an object and dividing the total by the object's area. These average values determine our binary threshold $T$. The whole process can be formulated as:
\begin{equation}
	\begin{split}
		T=  \frac{1}{n\times \lambda } { \sum_{i=1}^{n}} \frac{{\rm Sum} \big(\Theta  ({\rm Q}_{i})\big)}{\sqrt{{\rm size}({\rm Q}_i)}},
	\end{split}
\end{equation}
where ${\rm Q}_{i}$ is the $i$-th object proposal; $n$ is the total number of object proposals, and $\sqrt{{\rm size}(\cdot)}$ is the spatial size of ${\rm Q}_{i}$; $\rm Sum$ denotes the summation of the values of a gray map; $\textstyle \sum$ represents the summation operation of the scores of all the proposals; $\Theta$ mean the graying operation.
Specifically, $\lambda$ serves as a weight ranging from 0-1 to ensure the best performance of other methods.

\begin{figure*}[!t]
	\centering
	\includegraphics[width=0.82\linewidth]{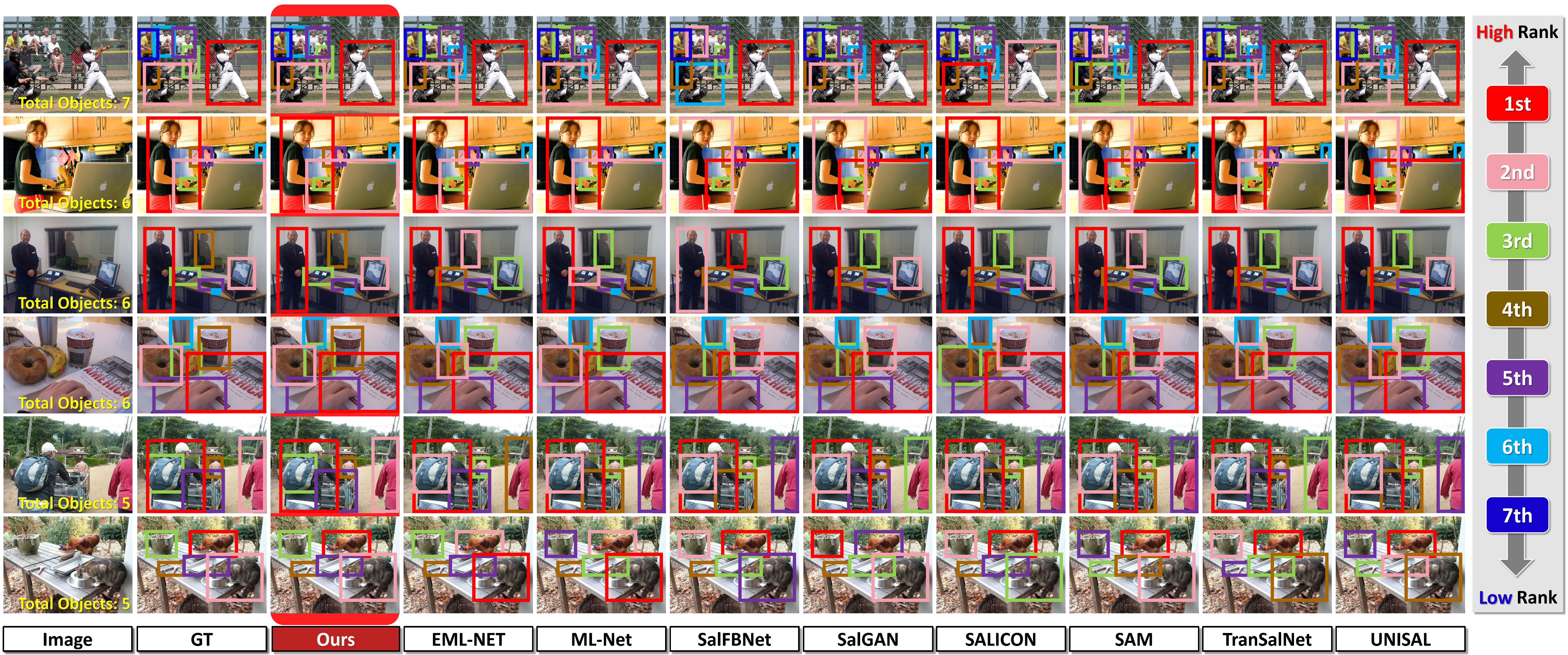}
	\vspace{-0.3cm}
	\caption{Visulizations of our proposed method and several competing methods. Zoom in for better observation.}
	\label{vis}
	\vspace{-0.5cm}
\end{figure*}

We present the quantitative comparison results for the SALICON dataset regarding the SRCC and F1 metrics in Fig.\ref{fig:SROCC}. We assign the saliency ranking orders of the compared methods using our newly proposed binary threshold $T$. Specifically, we set the GT threshold $\gamma$~(eq.~\ref{eq:gammabeta}) as 0.1, 0.2, 0.5, and 0.8 (we will detail the rationale of this choice in Fig.~\ref{fig:userstudy}-B later).
Our classification-based method, denoted as Ours (Cls.), outperforms all other methods regarding the SRCC metric on any GT/binary threshold, even when the best results of the compared methods are selected by the dynamic binary thresholds searching. Importantly, the SRCC metric of our method (0.907, 0.908, 0.905, and 0.896) remains consistent across all different binary value thresholds, indicating the robustness of our proposed method. Furthermore, our method performs competitively with other methods in terms of the F1. More qualitative visual comparisons with the competing methods are presented in Fig.~\ref{vis}.

In addition to our classification-based method, we also evaluated a regression-based version of our approach (referred to as "Ours (Reg.)"). However, the results demonstrate that the regression-based method is inferior to the classification-based method in all evaluation metrics. Our proposed classification-based method achieves average performance gains of 1.8\% and 4\% across four GT thresholds, as measured by the SRCC and F1 metrics. These results confirm the effectiveness of our classification-based approach.



\begin{table}[!t]
	\centering
	\caption{Quantitative comparisons. For a fair comparison, all compared three representative SOTA methods have been retained respectively using the default training split of the three tested sets.}
	\vspace{-0.3cm}
	\includegraphics[width=0.7\linewidth]{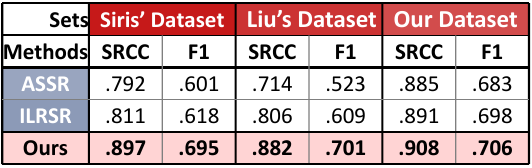}
	\label{tab:CompSOR}
	\vspace{-0.5cm}
\end{table}

\subsubsection{Comparison with Saliency Ranking Methods}
\label{CompSOR}

We compare our proposed model with two other representative SOTA saliency ranking methods, \emph{i.e.}, ASSR~\cite{Siris-CVPR20-raning} and ILRSR~\cite{Liu-TPAMI21-GCN}. For more comprehensive evaluations, we compare three models on three datasets, \emph{i.e.}, Siris’ dataset~\cite{Siris-CVPR20-raning}, Liu’ dataset~\cite{Liu-TPAMI21-GCN}, and our
proposed dataset. Following \cite{Liu-TPAMI21-GCN}, we only
select at most five instances in our prediction results for fair
comparisons due to the limited salient instances in the GT saliency map of the ASSR’ dataset. Other training settings are the same as \cite{Liu-TPAMI21-GCN}. Specifically, our model (we choose the best model with $\gamma$=0.2) and the compared two models are sorted differently, \emph{i.e.}, our model ranks the salient object orders in the proposal level (use rectangles to frame objects), while the two in pixel-wise level (segment the instances). Thus, we only compare the numerical ranking orders with SRCC and F1.

As shown in Table~\ref{tab:CompSOR}, we can observe that our
model generally outperforms the other two methods by a large
margin on all three datasets, \emph{e.g.}, compared with ILRSR, our model achieves an average performance gain of 8.6\%, 7.6\%, and 1.7\% in the SRCC metric across the three datasets.
These results show the effectiveness of our proposed saliency ranking model and its superiority and efficiency for practical usage.

\begin{table}[!t]
	\centering
	\caption{Components evaluation regarding the major parts of feature embedding \& enhancing (FEE, Sec.~\ref{sec:SPAFE})): a-c, PE; and HA (Eq.~\ref{eq:HA})) of proposed methods on our relabelled SALICON set.}
	\vspace{-0.3cm}
	\includegraphics[width=0.8\linewidth]{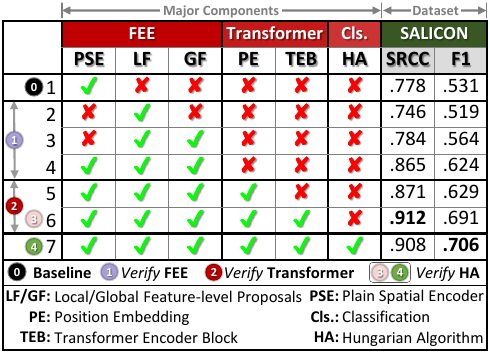}
	\vspace{-0.5cm}
	\label{tab:component}
\end{table}

\vspace{-0.4cm}
\subsection{Component Evaluation}
To validate the effectiveness of our method, we conducted an extensive component evaluation on our relabelled SALICON set. The results are shown in Table~\ref{tab:component}. To enable successful code running, we replaced the key components that needed to be verified with simpler operations. For example, we replaced the proposed SPAFE with simple ResNet101 and the Transformer with two fully-connected layers. We treated this replaced model as a baseline, and the qualitative result is shown in the 1st column denoted by mark {\large{\textcircled{\small{0}}}}.

Comparing line 1 and line 2, we can see that the features extracted by ResNet101 are more effective than the local feature-level features of the proposals obtained by ROIAlign, as evidenced by the higher SRCC and F1 metrics (0.778 \emph{v.s.} 0.746 and 0.531 \emph{v.s.} 0.519, respectively). This is likely because ResNet101's deep architecture makes its features more informative than the limited feature extraction capability of UNISAL, a lightweight network.


Lines 2-4 show the effectiveness of SPAFE in combining local and global feature-level proposals with ResNet101. Removing ResNet101 (line 3) or both ResNet101 and global feature-level proposals (line 2) decreased the SRCC metric in the SALICON set (mark {\large{\textcircled{\small{1}}}}). Global feature-level proposals enhance perception ability by capturing more surrounding features. ResNet101 provides more feature representations for Transformer blocks. Position embedding and Transformer Encoder Blocks are effective in capturing long-range dependency features (mark {\large{\textcircled{\small{2}}}}). We also conducted an ablation study on other backbones, such as FC and GCN (Table~\ref{tab:ShuffledGTMultiSingle}-C-a).

The Hungarian algorithm was found to be necessary in addressing the issue of identical classification results, as shown in marks {\large{\textcircled{\small{3}}}} and {\large{\textcircled{\small{4}}}}. Replacing the Hungarian algorithm with the max function resulted in decreased ranking accuracy (SRCC) despite improved F1 metric performance. This is because false-alarm ranking orders caused by the identical classification result issue may reduce ranking accuracy, which does not affect the SRCC metric.

\begin{figure*}[!t]
	\centering
	\includegraphics[width=0.85\linewidth]{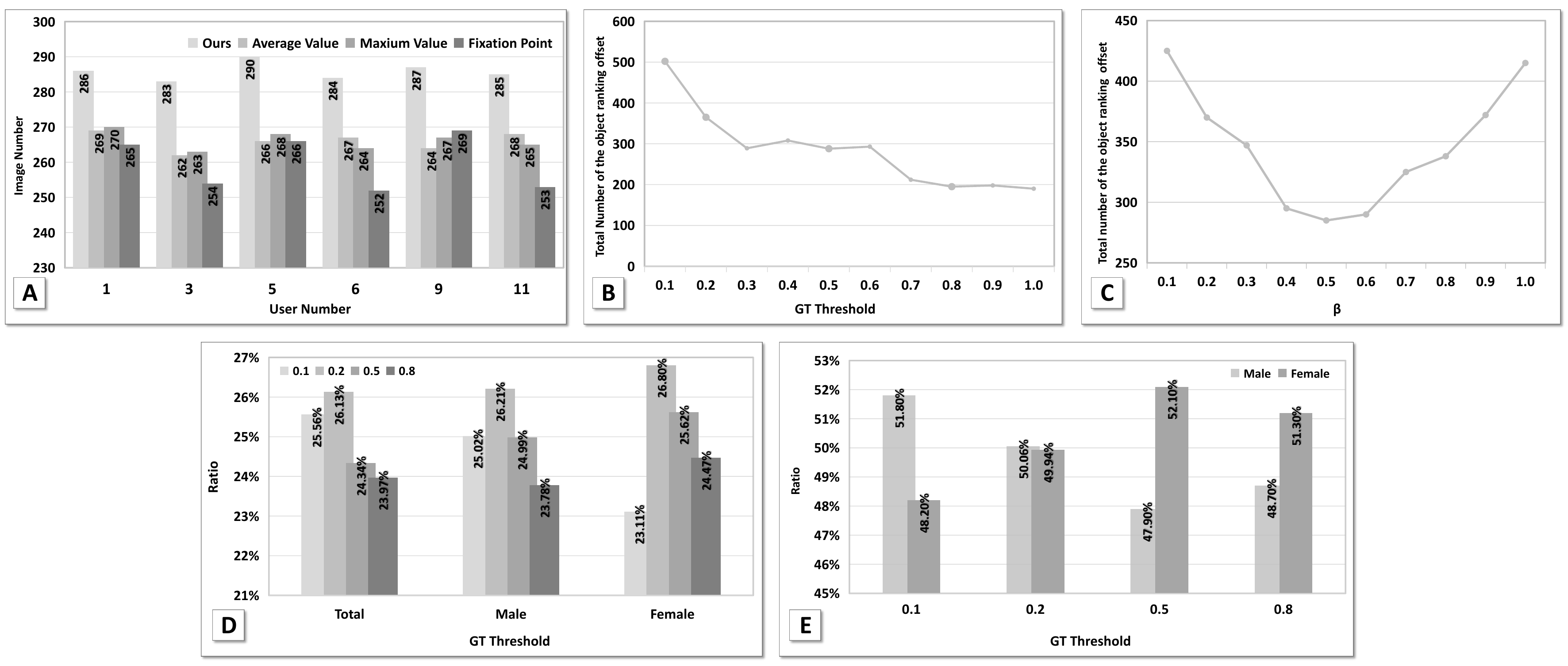}
	\vspace{-0.4cm}
	\caption{A: a user study on the consistency between real relative saliency ranking of the HSV and our saliency ranking GT orders, and other saliency ranking GT orders, respectively. B: a user study on discrepancy rate. That is to compute the total number of saliency ranking discrepancies ${\rm T}_{\rm offset}^{t}$ between adjacent thresholds $t-0.1$ and $t$. This involved summing the ranking changes of objects in all images under adjacent thresholds. C: ablation study on different choices of $\beta$. D: a user study on discrepancy rate. That is to compute the total number of saliency ranking discrepancies ${\rm T}_{\rm offset}^{t}$ between adjacent thresholds $t-0.1$ and $t$. This involved summing the ranking changes of objects in all images under adjacent thresholds. E: ablation study on different choices of $\beta$. }
	\label{fig:userstudy}
	\vspace{-0.5cm}
\end{figure*}

\vspace{-0.25cm}
\subsection{Our RA-SRGT \emph{v.s.} SOTA Competitors}
\label{userstudy}

\subsubsection{Effectiveness and rationality of RA-SRGT}
Saliency ranking is a subjective task. We aimed to eliminate induction bias by comparing four different saliency ranking GT orders generation methods to explore which one is closer to the real relative saliency of the human visual system.. These methods included ``our proposed method'', ``average value'', ``maximum value'', and ``fixation points''. Thus, we conducted a user study to validate our RA-SRGT~(Sec.~\ref{sec:RA-SRGT}).
We selected 300 images, each with four distinct saliency ranking GT orders generated by these four methods, and asked subjects to annotate the saliency ranking orders. We then calculated the number of images with the same ranking orders between annotated saliency ranking orders and saliency ranking orders generated by each of the four saliency ranking GT orders generation methods.
As shown in Fig.~\ref{fig:userstudy}-A, the number of images with the same saliency ranking orders between our method (the lightest bar) and the annotated real relative saliency of the human visual system steadily outnumbered the other methods as the number of subjects increased. This result demonstrates that our method is more feasible and closer to the human visual system and effectively generates accurate saliency ranking orders.

\subsubsection{Physiological exploration of RA-SRGT}
To conduct an in-depth analysis of saliency ranking GT orders from a physiological perspective, we performed a thorough investigation to determine the saliency ranking GT orders that best match the human visual system and are preferred by different individuals, using various thresholds. Specifically, we randomly selected 3,000 images from the SALICON dataset and generated saliency ranking GT orders by applying GT thresholds~(eq.~\ref{eq:gammabeta}) ranging from 0.1 to 1.0. We then computed the total number of saliency ranking discrepancy offsets ${\rm T}_{\rm offset}^{t}$ between adjacent thresholds $t-0.1$ and $t$. This involved summing the saliency ranking order changes of objects in all images under adjacent thresholds. It can be mathematically expressed as:
\begin{equation}
	{\rm T}^{t}_{\rm offset} = \sum_{n=1}^{N} \sum_{i=1}^{I} \left | {\rm Rank}_{n,i}^{t}-{\rm Rank}_{n,i}^{t-0.1} \right |,
\end{equation}
where $0.1\le t\le 1$ with a step size of 0.1; the index $i$ ($1\le i\le I$) and $I$ represents the $i$-th and total proposal(s) within the $n$-th image ($1\le n\le N$, $N$ = 3,000),  respectively; $\sum$ denotes summation operation; $\rm Rank$ ($1\le {\rm Rank}\le I$) is the saliency ranking order.

As shown in Fig.~\ref{fig:userstudy}-B, we observed that the total number of saliency ranking discrepancy offsets among thresholds from 0.3 to 0.6 and 0.7 to 1.0 are essentially the same. Therefore, we chose the transitional end-point values as our final GT thresholds $\gamma$~(eq.~\ref{eq:gammabeta}), namely, 0.1, 0.2, 0.5, and 0.8 (0.5 and 0.8 are the average thresholds of 0.3 to 0.6 and 0.7 to 1.0, respectively). These four GT thresholds could generate four different kinds of saliency ranking GT orders to the maximum extent. However, we did not know which one was preferred by the HVS and different individuals with diverse life backgrounds, \emph{e.g.}, gender. Therefore, we conducted another user study with over 100 subjects (aged 19 to 27) with diverse backgrounds to explore which of the four different kinds of saliency ranking GT orders of randomly selected 100 images under the four GT thresholds they preferred the most. We have also presented a quantitative comparison in Table~\ref{tab:ShuffledGTMultiSingle}-C-b.
\begin{figure}[!b]
	\centering
	\vspace{-0.6cm}
	\includegraphics[width=1.0\linewidth]{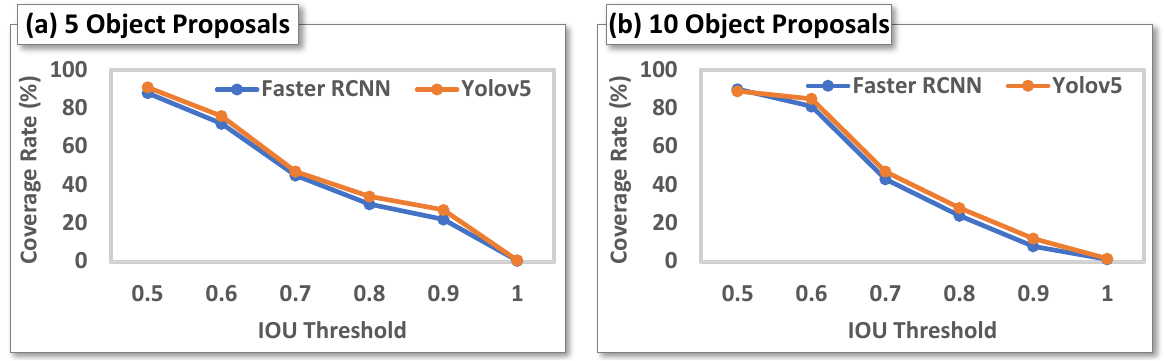}
	\vspace{-0.8cm}
	\caption{Charts of the changing trend of the proposal  coverage rates when replacing our object detector with four other object detectors, \emph{i.e.}, Faster RCNN~\cite{Faster-RCNN}, Yolov5~\cite{Yolov5}, DETR~\cite{DETR} and Sparse RCNN~\cite{Sparse-R-CNN}.}
	\label{fig:Robustness}
	\vspace{-0.1cm}
\end{figure}

Results in Fig.~\ref{fig:userstudy}-D showed that both males and females preferred the saliency ranking GT orders under GT threshold 0.2, which conforms to the quantitative comparison in Fig.~\ref{fig:SROCC}, where the SRCC and F1 metrics also obtained the best results under GT threshold 0.2. This finding verifies the effectiveness of our approach in generating saliency ranking GT orders, as our saliency ranking GT orders are closer to the real HVS. At the same time, we further explored the influence of gender in saliency ranking, as shown in Fig.~\ref{fig:userstudy}-E, where males were more inclined to saliency ranking GT orders under GT threshold 0.1, while females tended to choose GT threshold 0.5 and 0.8. Nevertheless, GT threshold 0.2, which both males and females preferred, was equal, demonstrating our approach's robustness.

\begin{table*}[!t]
	\centering
	\caption{Comparisons between models trained by ``Real GT'' and ``Shuffled GT'' (A), between multi-task learning and single-task learning (B), and between different choices of backbones and GT threshold $\gamma$ (C) on the SALICON set.}
	\vspace{-0.4cm}
	\includegraphics[width=1.0\linewidth]{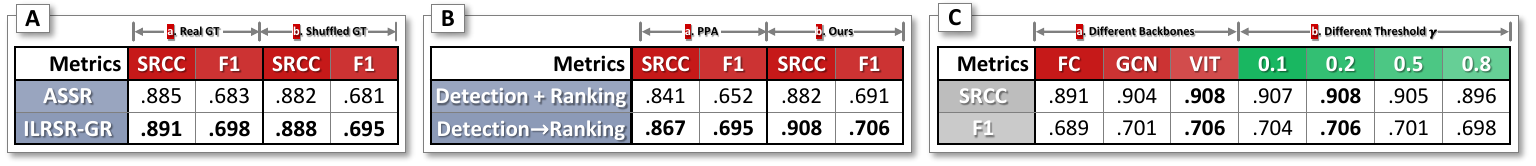}
	\label{tab:ShuffledGTMultiSingle}
	\vspace{-0.9cm}
\end{table*}

\subsubsection{Impact of the hyperparameter $\beta$}
To verify the impact of the hyperparameter $\beta$ in eq.~\ref{eq:gammabeta} on the generation of saliency ranking GT orders, we conduct an ablation analysis using the saliency ranking orders obtained from the user study, which includes 300 manually annotated images. To control variables, we fix the threshold $\gamma$ at 0.2. Next, we calculate the number of image offsets between the saliency ranking orders generated by our RA-SRGT (Sec.~\ref{sec:RA-SRGT}) and the saliency ranking orders manually annotated by humans under different $\beta$ values. As shown in Fig.~\ref{fig:userstudy}-C, the image offsets vary greatly with an increase in $\beta$, but remain almost constant in the range of 0.4$\sim$0.6. Results showed that when $\beta$ was too small, saliency ranking focused only on local fixations. When $\beta$ was in the range of 0.4$\sim$0.6, the ranking was more stable, and a large $\beta$ could ignore local objects. We set $\beta$ to 0.5 to balance the contribution of fixations and object spatial size.


\vspace{-0.3cm}
\subsection{Our Single-task Classification-based Method \emph{v.s.} SOTA}
\label{multi-task}

\subsubsection{Irrationality of instance segmentation-based saliency ranking methods}
To verify whether combining saliency ranking and instance segmentation tasks for multi-task learning~\cite{Liu-TPAMI21-GCN} may not necessarily improve model performance, we shuffled the ground truth segmentation mask ranking order \emph{i.e.}, shuffling the segmentation of each object in the segmentation mask by their brightness orders, and used it to retrain the existing saliency ranking model to test whether the model relies on the semantic information in the masks or not. We expected this would eliminate the side effect of segmentation and only leave the multi-task effect. However, as shown in Table~\ref{tab:ShuffledGTMultiSingle}-A, we found that the performance of original models trained by ``Shuffled GT'' stays the same as the original models (``Real GT''). This suggests that the segmentation masks with consistent brightness orders and saliency ranking orders of each object do not significantly influence the saliency ranking task, which means that the segmentation task only provides segmented instance results for the saliency ranking task and has not much in the way of performance improvement. Thus, it might not need instance segmentation as an auxiliary task for saliency ranking.

\subsubsection{Effectiveness of single-task method with offline proposal generation}
To demonstrate the drawbacks of the existing multi-task model~\cite{Fang-ICCV21-PPA} that uses online object proposal generation and training of the detector for saliency ranking task, we design a single-task model that uses offline object proposal extraction and no training of the detector. The hypothesis is that the online extraction and training of the detector might introduce noise or inconsistency in the object features and masks, which may affect the ranking performance. The single-task model avoids this problem by using a fixed and reliable object detector pre-trained on a large-scale dataset. We compare the two models \emph{i.e.}, 1) original model~\cite{Fang-ICCV21-PPA} with multi-task learning of two parallel object detection branch and saliency ranking branch, and 2) modified model~\cite{Fang-ICCV21-PPA} with single-task learning by offline generated object proposals and an online saliency ranking on SALICON dataset. Experimental results in Table~\ref{tab:ShuffledGTMultiSingle}-B show that the single-task model will achieve comparable or better performance than the multi-task model, while being more efficient and stable.

\vspace{-0.3cm}
\subsection{In-depth Analysis of Multi-task Learning-based Saliency Ranking Methods}
\label{issue2}
\subsubsection{High Complexity}
In Fig.~\ref{fig:motivation1}-B (a), we discussed the high complexity of existing multi-task learning methods that use segmentation as a basis. These methods are more intricate than plain object detection-based approaches. As saliency ranking is designed to determine the visual significance of objects/regions in an image, without needing pixel-level segmentation or salient detection of the entire image, it simplifies the saliency ranking tasks and improves efficiency.
To evaluate the computational efficiency of our proposed method, we conducted a comparison of the running time between our method and several segmentation-based state-of-the-art (SOTA) methods, including ASSR~\cite{Siris-CVPR20-raning}, ILRSR~\cite{Liu-TPAMI21-GCN}, PPA~\cite{Fang-ICCV21-PPA}, RSDNet~\cite{Kalash-TPAMI21}, and Bi-OCPL~\cite{Tian-CVPR22-BI}. As shown in Table~\ref{time0}-A, our method achieved real-time speed, running at 20.5 frames per second (FPS) during the \textbf{total inference phase}, including the offline generation of proposals.

\begin{table}[!t]
	\centering
	\caption{Comparisons of computational efficiency and resource requirements between our proposed and other competing methods. FPS is considered to evaluate the model efficiency. \textbf{Bold} indicates the best performance.}
	\vspace{-0.2cm}
	\includegraphics[width=0.85\linewidth]{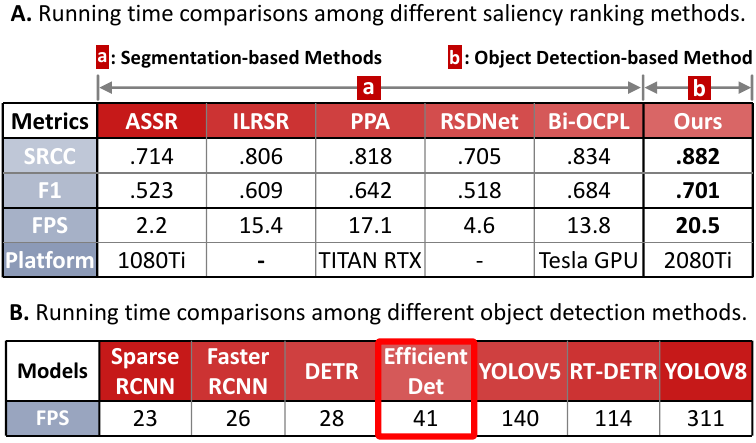}
	\vspace{-0.4cm}
	\label{time0}
\end{table}

Furthermore, the computational cost of offline object proposal generation depends on the selected object detection models. We present several representative SOTA object detection models in Table~\ref{time0}-B. Although the FPS of EfficientDet - the model chosen in our method - lies in the middle, it still outperforms other models (Table~\ref{time0}-A).



\begin{table}[!t]
	\centering
	\caption{Comparisons among multi-task-learning segmentation/detection-based saliency ranking models (a and b) and our single-task-learning model (c). The metric ``mIoU'' is used to evaluate the performance of segmentation models, while the metrics ``AP/AP$_{50}$'' are employed for assessing the performance of detection models.}
	\vspace{-0.2cm}
	\includegraphics[width=0.85\linewidth]{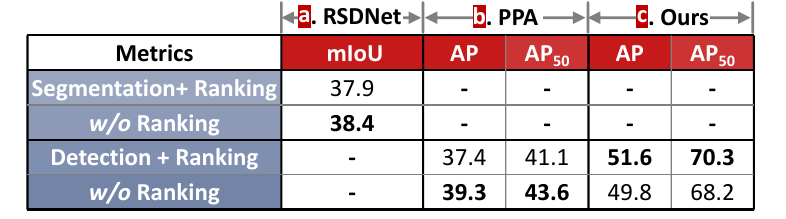}
	\vspace{-0.5cm}
	\label{UncertainMutualPromotion}
\end{table}

\subsubsection{Uncertain Mutual Promotion}
In multi-task learning, ``mutual promotion'' is a strategy or approach aimed at improving the performance of multiple related tasks by promoting mutual interaction and sharing of information between the tasks.
This implies that shared features and knowledge between tasks can mutually promote and benefit all tasks.
However, in practice, ``mutual promotion'' may encounter some issues because the relationships between different tasks are not always clear or certain. This gives rise to the concept of ``uncertain mutual promotion'', which emphasizes the uncertainty in the relationships between tasks, where certain tasks' features may have positive impacts on other tasks. In contrast, others may have no effect or even negative effects.

To demonstrate the phenomenon of ``uncertain mutual promotion'' in existing saliency ranking models, we conducted two validations using a multi-task learning strategy. Specifically, 
1) we tested the performance of only one of the tasks in multi-task learning, \emph{e.g.}, the performance of segmentation in segmentation-based saliency ranking models or the performance of object detection in object detection-based saliency ranking models\footnote{Note that nearly all the existing saliency ranking models are segmentation-based, only a few models resort to object detection before segmenting the objects, we consider these kinds of methods containing object detection methods as object detection-based saliency ranking models.}; 
2) we trained and tested only one task by excluding the saliency ranking task from multi-task learning, \emph{e.g.}, the segmentation task or object detection task. 
selected RSDNet~\cite{Kalash-TPAMI21} as the representative segmentation-based model, and PPA~\cite{Fang-ICCV21-PPA} as the object detection-based model due to its easy code availability.

Suppose segmentation-based multi-task-learning model \textbf{a} (RSDNet) can achieve mutual promotion in multi-task learning. In that case, the expected outcome should be that segmentation performance during simultaneous segmentation and ranking is higher than when only segmentation is performed. However, according to the results in Table~\ref{UncertainMutualPromotion}, the data decreased from 38.4 to 37.9 regarding the mIOU metric, which provides sufficient evidence that segmentation-based multi-task-learning model \textbf{a} (RSDNet) is an example of uncertain mutual promotion. As is the same case in object detection-based multi-task-learning model \textbf{b} (PPA), the AP and AP$_{50}$ separately decreased, which also proves that object detection-based multi-task-learning model \textbf{b} (PPA) is also an example of uncertain mutual promotion. However, in our approach (model \textbf{c} (Ours)), a single-task learning-based saliency ranking model, the ranking task offers support to its pre-object detection processing, boosting the AP and AP$_{50}$ separately from 49.8 to 51.6 and 68.2 to 70.3. It proves that our single-task learning method can achieve mutual promotion.

The ``uncertain mutual promotion'' in segmentation-based multi-task learning models can be attributed to two factors. Firstly, there is task misalignment between the segmentation task and the ranking task. These tasks have different objectives and optimization criteria, which hinders their mutual promotion. Secondly, training dynamics play a role, as the optimization process may prioritize one task over the other, leading to decreased performance in the segmentation task.
In contrast, our approach avoids ``uncertain mutual promotion'' for several reasons. We use a single-task learning approach, focusing solely on saliency ranking without conflicting tasks. Additionally, we employ pre-object detection processing, ensuring accurate object delineation and providing valuable information for ranking. This targeted optimization and alignment of objectives enhance the ranking performance without compromising segmentation accuracy.

Based on this observation, we conclude that existing multi-task learning-based saliency ranking models can exhibit ``uncertain mutual promotion'' and our single-task learning-based saliency ranking method is ``mutual promotion''.

\begin{table}[!t]
	\centering
	\caption{The final ranking results comparisons among five object detection methods. The final ranking results exhibit only a small range of change.}
	\vspace{-0.2cm}
	\includegraphics[width=0.7\linewidth]{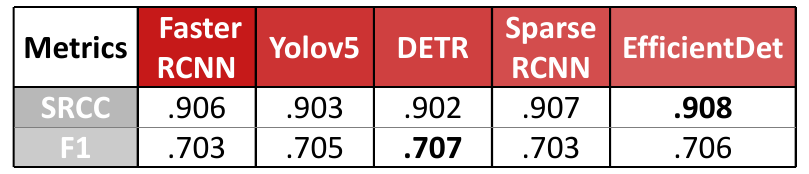}
	\vspace{-0.5cm}
	\label{stabletraining}
\end{table}

\subsubsection{Unstable Training}
We conducted experiments by replacing our original object detection methods with EfficientDet, Faster RCNN~\cite{Faster-RCNN}, Yolov5~\cite{Yolov5}, DETR~\cite{DETR}, and Sparse RCNN~\cite{Sparse-R-CNN}. The quantitative results presented in Table~\ref{stabletraining} demonstrate that the final ranking results only exhibit a small range of change. This validates that our approach possesses a strong capability for ``stable training''.

We argue that multi-task learning models exhibit ``unstable training'' due to two primary reasons: 1) In multi-task learning, the saliency ranking tasks require retraining the entire model whenever the segmentation model changes. In contrast, our model only needs to be trained once using high-accuracy, efficient object detection methods. The similarity of generated proposals from different detection methods allows us to directly feed offline-generated object proposals into the trained saliency ranking model without training a new model for each detection method.
2) Joint training of segmentation-based multi-task learning models can prioritize one task over the other, leading to an out-of-sync optimization process and ``unstable training''. Our single-task learning method focuses solely on the saliency ranking loss, simplifying implementation and avoiding resource allocation conflicts.

\vspace{-0.3cm}
\subsection{Method Generality Analysis}

\begin{figure}[!t]
	\centering
	\includegraphics[width=1\linewidth]{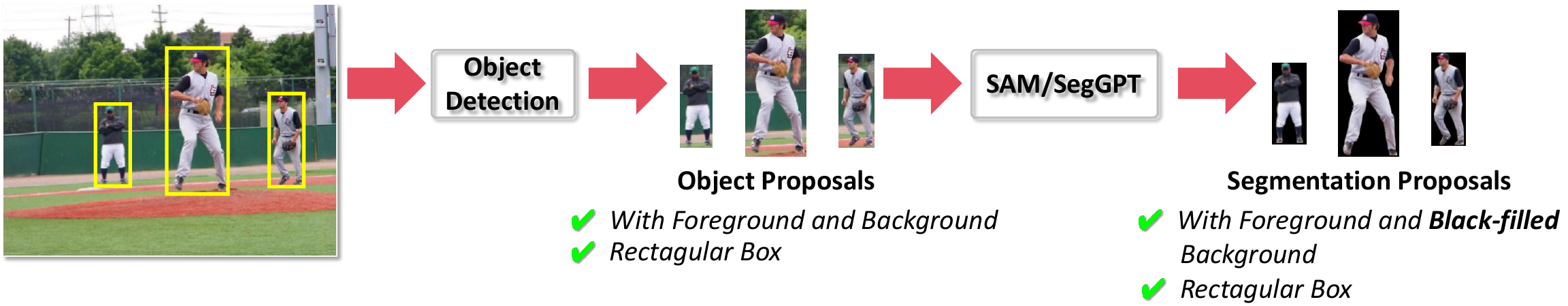}
	\vspace{-0.7cm}
	\caption{Comparisons between object proposals and segmentation proposals.}
	\label{generalitySAM}
	\vspace{-0.5cm}
\end{figure}

We evaluated our detection-based saliency ranking method's compatibility with segmentation tasks, specifically segmentation-based saliency ranking. We used large-scale segmentation models such as SAM~\cite{SAM} and SegGPT~\cite{SegGPT} to generate rectangular segmentation proposals, which were fed into our proposed saliency ranking model. These proposals differ from object proposals in that they only contain foreground objects (with the background filled in black) while also being rectangular in shape. We initially obtained object proposals using object detection models, which we then fed into SAM or SegGPT for segmentation (refer to Fig.~\ref{generalitySAM}).


The results in Table~\ref{generality} show that our object proposal-based method (referred to as Ours (Detection)) performs slightly lower compared to the segmentation proposal-based methods, namely Ours (SAM) and Ours (SegGPT). Furthermore, these segmentation proposal-based methods outperform other segmentation-based saliency ranking models listed in Table~\ref{time0}. These segmentation proposal-based methods demonstrate higher performance because they only focus on ranking the objects, while our object proposals include unnecessary background information, which may impact the model's learning process.
However, it is worth noting that the segmentation proposal-based methods have longer running times due to the complexity of pixel calculations involved in segmentation tasks, whereas the object proposal-based method deals with sparse tasks. This discrepancy arises from the nature of these two approaches.


To further explore the effectiveness of segmentation models in saliency ranking, we used large-scale models like SAM and SegGPT to generate masks. These masks were then used to fill proposals based on object sizes. Results in Table~\ref{mask2box} show that using SAM and SegGPT for mask-to-proposal transformation improves saliency ranking performance. However, their running speed (16.8 and 17.3 FPS) is slower compared to our object detection-based model (20.5 FPS). This is because SAM and SegGPT perform pixel-wise segmentation across the entire image, resulting in more accurate object localization. Despite the slight improvement in ranking performance, these methods reduce processing speed. Considering the goal of saliency ranking is to assign a saliency order to each object, object detection-based methods strike a balance between effectiveness and efficiency. These findings validate the strong stability and generality of our proposed method.

\begin{table}[!t]
	\centering
	\caption{Generality analysis of our detection-based saliency ranking method to segmentation-based tasks.}
	\vspace{-0.2cm}
	\includegraphics[width=0.7\linewidth]{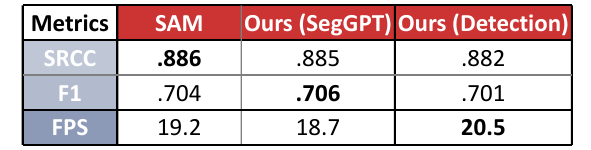}
	\vspace{-0.35cm}
	\label{generality}
\end{table}	

\begin{table}[!t]
	\centering
	\caption{Comparisons between transforming segmentation masks to object proposals and our directly generated object proposals.}
	\vspace{-0.2cm}
	\includegraphics[width=0.9\linewidth]{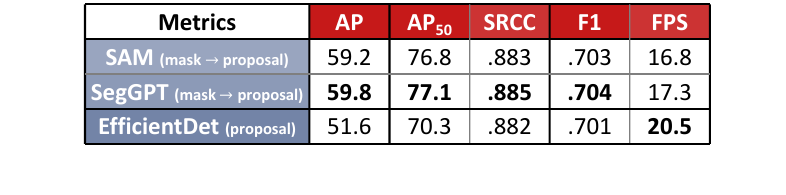}
	\label{mask2box}
	\vspace{-0.9cm}
\end{table}

\vspace{-0.25cm}
\subsection{Ablation Study}
\subsubsection{Robustness of our object-wise proposal detection}
\label{Robustness}

We evaluated the robustness of our proposal detection method by comparing the coverage rates of proposals detected by different object detectors. We used EfficientDet as the baseline and compared it with Faster RCNN~\cite{Faster-RCNN}, Yolov5~\cite{Yolov5}, DETR~\cite{DETR} and Sparse RCNN~\cite{Sparse-R-CNN}. Varying the IOU threshold from 0.5 to 1, we observed consistent coverage rates for proposals detected by our used detector (\emph{e.g.}, EfficientDet) and other detectors (\emph{e.g.}, Faster RCNN) (see Fig.~\ref{fig:Robustness}). This indicates the robustness of our approach across scenarios with 5 (a) and 10 (b) proposals. Our training approach ensured stable performance across various detectors, demonstrating strong generalizability and applicability of our proposed method.

\subsubsection{Transformer v.s. other backbones}

We conducted an ablation study on the Vision Transformer (VIT), Graph Convolutional Network (GCN), and fully-connected layer (FC) for saliency ranking. The object proposals determined GCN's number of graph nodes, while FC directly received the salient position-aware feature $F_{p}$ (Sec.~III-C2) as input for predicting the final results.
The results in Table \ref{tab:ShuffledGTMultiSingle}-C-a indicate that VIT outperformed GCN and FC in modeling object relationships, scoring 0.908 compared to 0.904 for GCN and 0.891 for FC. VIT's self-attention mechanisms enabled it to capture long-range dependencies and understand spatial relationships between objects comprehensively. GCN faced challenges in achieving this efficiently through graph-based techniques. FC directly input the salient feature without considering spatial relationships, resulting in information loss and inferior performance compared to VIT and GCN.



\subsubsection{Impact of GT threshold $\gamma$}
To further verify the advantages of our proposed relationship-aware saliency ranking GT orders generation method, we conducted a quantitative test to explore the impact of GT threshold $\gamma$ (eq.~4), where we chose $\gamma$ = 0.1, 0.2, 0.5, and 0.8. Results in Table~\ref{tab:ShuffledGTMultiSingle}-C-b show that as the $\gamma$ value increases, the SRCC and F1 metrics first rise and then fall simultaneously. For instance, when the $\gamma$ value changes from 0.1 to 1.0, the SRCC and F1 metrics decline by 1.1\% and 0.6\%. However, when the $\gamma$ value is 0.2, the SRCC and F1 metrics reach their maximum. This is because when $\gamma$ is small, the spatial size of the object has less impact on the final saliency ranking GTs, whereas when $\gamma$ is large, the object size has too much influence, resulting in false alarms. These results confirm the robustness of our proposed method compared to other saliency ranking GT order generation methods.

\vspace{-0.2cm}
\subsection{Limitations}

\begin{figure}[!t]
	\centering
	\includegraphics[width=0.75\linewidth]{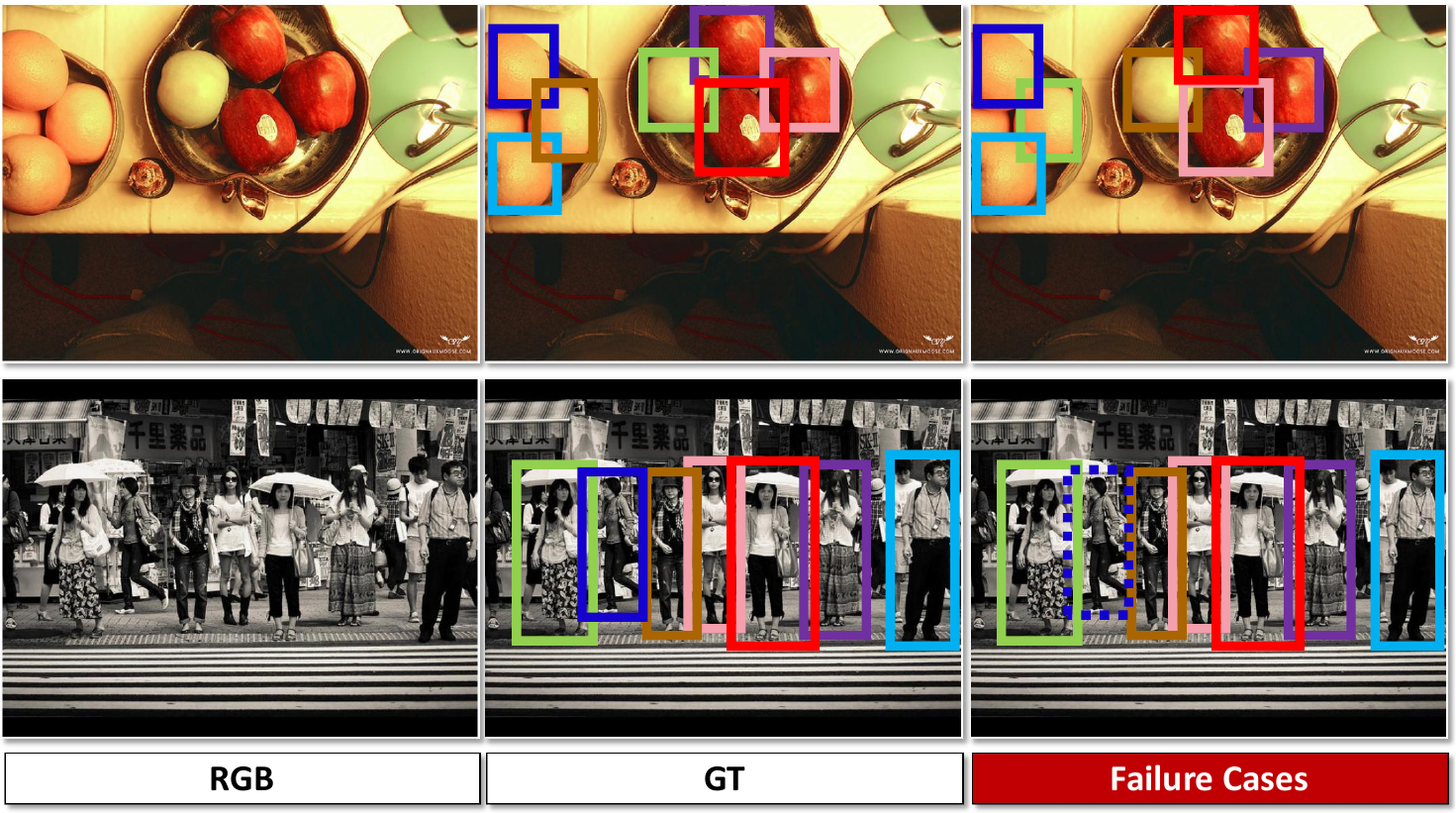}
	\vspace{-0.3cm}
	\caption{Failure cases with similar appearances and shapes (1st line) and object detector failing to detect all object proposals (2nd line).}
	\label{fig:failure}
	\vspace{-0.5cm}
\end{figure}

1) In Fig.~\ref{fig:failure}, we demonstrate the limitations of our model through visual examples. These cases highlight challenges in distinguishing visually similar objects with comparable saliency ranking orders (line 1). To address this, we can incorporate color saliency and leverage existing fixation prediction models to locally rank visual stimuli based on color and shape. Another challenge arises when our model fails to detect all object proposals, making it difficult to assign saliency ranking orders (line 2). To mitigate this, we can utilize a more accurate object detection model or employ multiple detection models in coordination. Additionally, treating the problem of missing detections as a camouflaged object detection task~\cite{Ji2,Ji3} could be a straightforward approach. Camouflaged object detection methods are specifically designed to handle scenarios where objects blend with the background. While this problem differs from saliency detection~\cite{Ji1}, we will not explore it further in this discussion.


2) Our RA-SRGT method (Sec.~\ref{sec:RA-SRGT}) has several limitations that require attention. Firstly, the approach relies heavily on object proposals generated by detection models, which may introduce errors in object localization, impacting saliency ranking accuracy. It is crucial to find ways to mitigate these errors.
Secondly, the method's thresholds, $\gamma$ and $\beta$, are manually set, introducing subjectivity. This subjectivity poses challenges in generalizability across datasets or scenarios where optimal thresholds may vary.


\section{Conclusion}

In summary, this paper introduces a novel paradigm for saliency ranking that outperforms existing methods on the SALICON dataset. The approach addresses challenges in generating ground truth orders, network design, and training protocols. A key contribution is recognizing the limitations of existing saliency detection methods in determining the relative importance of multiple objects and their relationships. The proposed whole-flow processing paradigm demonstrates superior performance, accurately ranking salient objects and enhancing downstream tasks. These findings pave the way for improved saliency ranking methods and offer valuable insights into the importance ordering of visually salient objects.
Future work can focus on improving data efficiency through semi-supervised learning or transfer learning, and enhancing generalization capabilities for diverse datasets and real-world deployment.

\bibliographystyle{ieeetr}
\bibliography{egbib}
\end{document}